\documentclass[runningheads]{llncs}

\usepackage{eccv}

\usepackage{eccvabbrv}

\usepackage{graphicx}
\usepackage{booktabs}
\usepackage{threeparttable}
\usepackage{subcaption}
\usepackage{tabularx}
\usepackage{wrapfig}
\usepackage{siunitx}
\usepackage{amsfonts}
\usepackage{algorithm}
\usepackage{algpseudocode}
\usepackage{tikz}
\usetikzlibrary{arrows.meta, positioning, fit, backgrounds, calc}

\usepackage{pifont}
\usepackage{multirow}
\usepackage{colortbl}

\usepackage[accsupp]{axessibility}

\usepackage{hyperref}

\usepackage{orcidlink}

\begin{document}

\title{SAFER-Activities: A Dataset for Smart Assessment of Fall Events and Routine Activities}

\titlerunning{SAFER-Activities Dataset}

\author{Diwas Lamsal\inst{1}\thanks{Work done while at the Asian Institute of Technology.}\orcidlink{0009-0006-1125-2299} \and
Pramod Wickramatilake\inst{2}\orcidlink{0009-0005-2755-0998} \and
Jednipat Moonrinta\inst{2}\orcidlink{0000-0002-7918-007X} \and
Mongkol Ekpanyapong\inst{2}\orcidlink{0000-0002-0192-6249} \and
Matthew N. Dailey\inst{2}\orcidlink{0000-0002-7191-3558}}

\authorrunning{D.~Lamsal et al.}

\institute{KU Leuven, Leuven 3000, Belgium\\
\email{diwas.lamsal@kuleuven.be}
\and
Asian Institute of Technology, Khlong Luang, Pathum Thani 12120, Thailand}

\maketitle

\begin{abstract}
Smart healthcare monitoring systems require precise action recognition to ensure well-being and timely intervention in critical situations such as falls, particularly for mobility-challenged individuals. Existing datasets are often clip-based, lacking the frame-level detail needed to recognize actions online, as they unfold. To address this, we introduce SAFER-Activities, a dataset for fall detection and physical activity monitoring, with a dedicated subset for wheelchair use scenarios. It comprises over 66 hours of video data captured by multiple cameras, with 85,310 action instances and frame-level annotations for 30 action classes. We benchmark action recognition on SAFER-Activities with 2D and 3D skeleton models, RGB models with frozen backbones, and multimodal fusion strategies, and evaluate on in-lab, out-of-distribution, and cross-dataset test sets. Skeleton-based models generalize best under domain shift; fusing frozen RGB features with the skeleton stream improves in-domain recognition over the baseline CNN1D, most clearly on the wheelchair subset, but degrades out of distribution. Cross-dataset and qualitative evaluations confirm that models trained on SAFER-Activities transfer well to unseen environments and external fall data. To support research on robust fall detection and activity monitoring, we release the dataset and code at \url{https://safer-activities.github.io/}.
\end{abstract}

\section{Introduction}
\label{sec:intro}

Smart healthcare monitoring systems should be designed to analyze activities of daily living (ADL), ensuring that people maintain an adequate level of physical activity~\cite{whoguidelines}. They should also enable timely intervention in critical situations like falls, which are a leading cause of injury-related hospitalizations~\cite{who2008falldata}. Most smart healthcare monitoring systems are based on wearable sensors or cameras~\cite{leuvenfalldataset}. Wearable sensors can be uncomfortable for some users, and cognitively impaired individuals often forget to wear them~\cite{Fleminga2227fallcohortstudy}. Camera-based human action recognition (HAR) systems are effective complementary or standalone solutions.

HAR is a central task in video understanding, with comprehensive datasets contributing to its progress in the last decade \cite{Shahroudy2016,Kinetics}. Various features, including RGB \cite{quovadisactionrecognition, temporalsegmentnets, tong2022videomae}, optical flow \cite{twostreamconvnets, quovadisactionrecognition}, and human skeletons \cite{stgcnpaper, Duan2022, msg3dpaper}, have been explored for HAR. Skeleton-based methods offer concise representations of body movements that are robust to background clutter and lighting changes~\cite{Duan2022}. However, they discard scene context and can become unreliable when pose estimation is degraded by occlusion or unusual viewpoints~\cite{Li2019}. RGB-based methods capture richer semantic information, including scene context and object interactions, but are more susceptible to domain shift, as appearance features learned from one environment often fail to transfer to new settings \cite{datasetbias}. Combining both modalities is therefore an appealing direction~\cite{mmcl}.

\begin{figure}[tb]
  \centering
  \includegraphics[width=\linewidth]{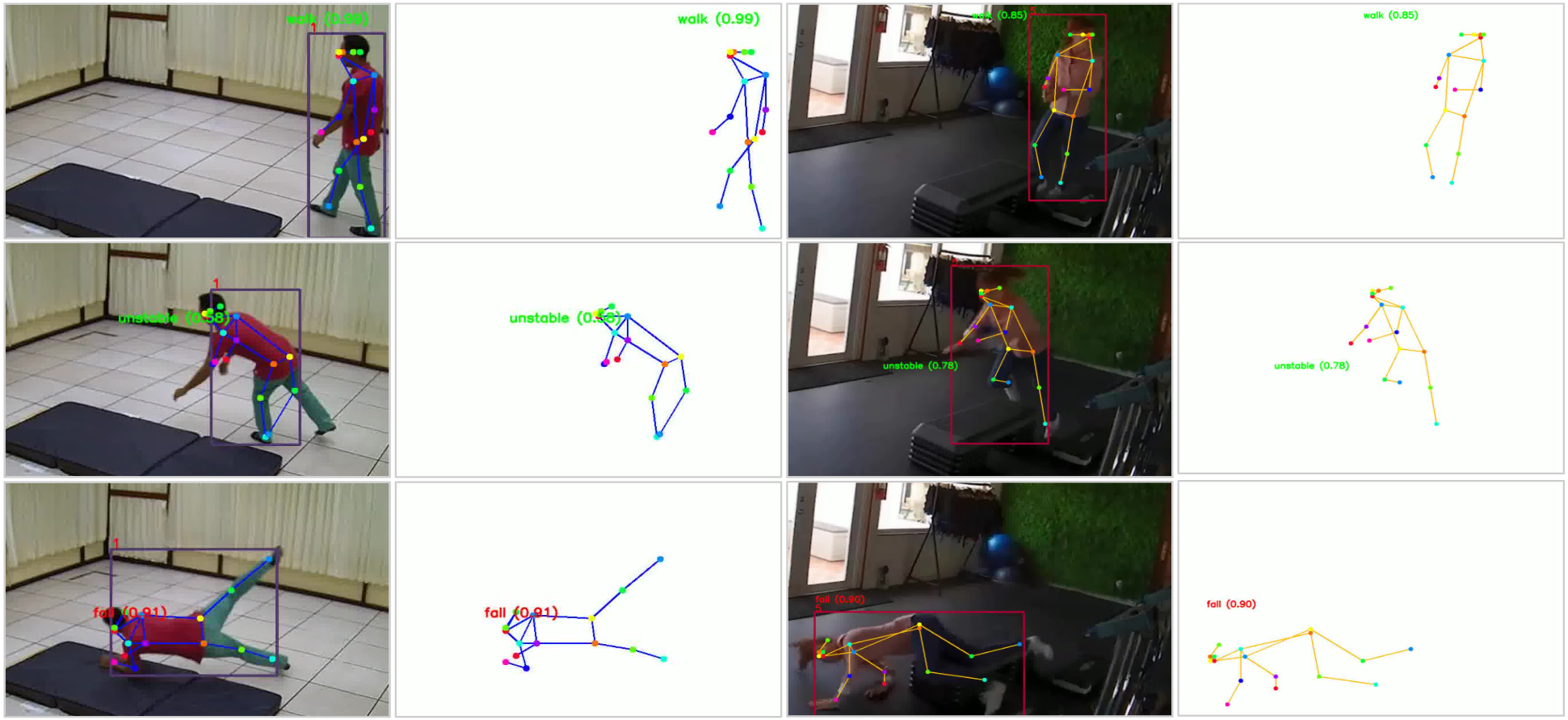}
    \caption{Predictions from a 2D skeleton-based model trained on SAFER-Activities, evaluated on unseen in-lab test data (left) and a real-world fall example (right).}
  \label{fig:skeleton_demonstration}
\end{figure}

HAR for smart healthcare monitoring requires datasets with long, untrimmed videos and precise frame-level annotations \cite{fineactionLiu2022} rather than the short, coarsely labeled clips typical of large-scale activity recognition benchmarks \cite{Shahroudy2016, Kinetics}. These datasets must include a broad range of routine activities, such as walking and exercising, to enable physical activity monitoring, as well as actions resembling falls, such as lying down on a sofa, to reliably distinguish them from actual falls. They should also include evaluation data from unseen subjects and environments to verify that trained models generalize beyond the controlled training domain. While prior fall detection datasets \cite{imviadataset, leuvenfalldataset} feature realistic falls designed to mimic real-world scenarios, they are limited by a small number of fall instances and lack comprehensive labeling of routine and fall-like events.

Another gap in the activity recognition resources currently available lies in the unavailability of datasets containing wheelchair users. This group is particularly vulnerable and would greatly benefit from physical activity monitoring. Prolonged wheelchair use, especially propelling and lifting, can cause shoulder damage \cite{detectionwheelchairactivities}. Wheelchair users also face heightened risks of falls during transfers \cite{wheelchairfalldata}. Insufficient physical activity by some wheelchair users puts them at risk of chronic diseases \cite{Booth2012}, hypertension, hyperlipidemia, and diabetes \cite{Popp2018}. Therefore, developing robust activity recognition systems for this population is crucial for promoting their health, safety, and overall well-being.

To address these issues, we introduce SAFER-Activities: a large-scale dataset for fall detection and activity monitoring, with over 66 hours of video data from 46 participants, with frame-level annotations for 85,310 action instances across 30 classes, including 5,406 fall instances alongside routine and closely related actions, such as lying down and sitting. The dataset includes a subset focused on wheelchair use and a separate non-lab test set recorded in a home environment to evaluate real-world generalization. We also provide a complementary pose estimation dataset of 6,846 images and 8,324 human instances to benchmark pose estimation methods for people seated in wheelchairs.

We benchmark SAFER-Activities with a set of methods spanning 2D and 3D skeleton-based models, RGB-based models using pretrained frozen visual backbones, and multimodal fusion. Our evaluation covers in-lab, non-lab, and wheelchair test sets, as well as cross-dataset generalization on an external fall detection dataset. We find that while skeleton-based methods offer the strongest out-of-distribution generalization, multimodal fusion improves in-distribution performance but introduces challenges under domain shift, highlighting SAFER-Activities as a valuable testbed for robust action recognition and fall detection research. Models trained on SAFER-Activities effectively detect falls in unseen real-world scenarios (\cref{fig:skeleton_demonstration}). Our contributions are as follows:

\begin{itemize}
    \item We introduce SAFER-Activities, a large-scale dataset with dense frame-level annotations for falls and routine activities, multi-camera viewpoints, and a dedicated subset with wheelchair use scenarios, serving as a comprehensive benchmark for action recognition and fall detection.
    \item We provide extensive benchmarks spanning 2D pose, 3D pose via monocular lifting, frozen RGB-based pretrained backbones, and multimodal fusion methods across in-lab, out-of-distribution, and wheelchair evaluation splits.
    \item We include a non-lab test set recorded in a home environment with unseen participants and viewpoints, and perform cross-dataset evaluation on an external fall detection dataset.
    \item We release a complementary dataset to benchmark human pose estimation methods during wheelchair use.
\end{itemize}

\section{Related Work}

\subsection{Fall Detection Datasets}

\begin{table}[tb]
  \centering
\caption{Comparison of vision-based fall detection datasets.
    \#~Subj.: number of subjects.
    Age: age range of subjects.
    \#~Classes: total number of labeled activity categories
    (2~indicates fall/non-fall labels only).
    \#~Falls: total fall instances across all camera views.
    \#~ADL Inst.: total labeled non-fall activity instances
    (\ding{55}~if ADL are not individually labeled).
    \#~WC Falls: wheelchair fall instances
    (\ding{55}~if not included).
    \#~Hours: total duration of video data.
    N/A: not available.
    Only Auvinet~\etal~\cite{multiplecamerasfalldataset} and SAFER-Activities provide frame-level annotations for all activity classes.}
  \label{tab:fall_dataset_comparison}
  \resizebox{\textwidth}{!}{
  \begin{tabular}{@{}lcccccccc@{}}
    \toprule
    Source & \# Subj. & Age & \# Classes & \# Falls & \# ADL Inst. & \# WC Falls & \# Hours \\
    \midrule
    Guerrero~\etal~\cite{caucafall} & 10 & 23--40 & 2 & 50 & \ding{55} & \ding{55} & $<$1 \\
    Charfi~\etal~\cite{imviadataset} & 9 & N/A & 2 & 143 & \ding{55} & \ding{55} & $<$1 \\
    Mart\'inez~\etal~\cite{upfalldataset} & 17 & 18--24 & 11 & 255 & 306 & \ding{55} & $\sim9.4$ \\
    Auvinet~\etal~\cite{multiplecamerasfalldataset} & 1 & N/A & 9 & 200 & 1120 & \ding{55} & $\sim10.6$ \\
    Baldewijns~\etal~\cite{leuvenfalldataset} & 10 & N/A & 2 & 275 & \ding{55} & 20 & $\sim41.4$ \\
    \midrule
    \textbf{SAFER-Activities} & \textbf{46} & \textbf{18--58} & \textbf{30} & \textbf{5,406} & \textbf{79,904} & \textbf{840} & \textbf{$>$66} \\
    \bottomrule
  \end{tabular}
  }
\end{table}

Several small-scale datasets exist for vision-based fall detection (\cref{tab:fall_dataset_comparison}). Guerrero~\etal~\cite{caucafall} collect 50 fall instances from 10 participants in an uncontrolled environment with varying lighting, but label only fall versus non-fall. The ImViA dataset~\cite{imviadataset} features 222 videos with 143 fall instances across four realistic indoor settings; however, only fall boundaries are annotated. Mart\'inez~\etal~\cite{upfalldataset} present UP-Fall, a multimodal dataset combining cameras with wearable and ambient sensors, providing clip-level annotations for 11 activity classes across 17 participants, all young adults aged 18--24. Auvinet~\etal~\cite{multiplecamerasfalldataset} provide 24 scenarios captured by eight cameras with frame-level labels for 9 activity categories and 1,120 ADL instances, but feature only a single participant. Baldewijns~\etal~\cite{leuvenfalldataset} re-enact actual nursing-home falls across 72 scenarios from five camera angles, including 20 wheelchair fall instances, and emphasize realism with longer videos containing post-fall activities, but do not provide ADL labels.

Relative to prior vision-based fall-detection datasets, SAFER-Activities provides substantially more fall instances (5,406), frame-level annotations for 30 activity classes, and recordings from 46 adult participants spanning a broader age range (18--58). It also includes diverse routine and fall-like activities that cover typical scenarios encountered in realistic environments. Among the compared datasets, the ImViA dataset~\cite{imviadataset} is particularly suited for evaluating cross-dataset generalization, as it features a realistic home environment with variable lighting and precise fall boundary annotations. We use it for cross-dataset evaluation (see \cref{tab:fall_accuracy_on_existing_datasets}).

\subsection{Camera-Based Human Action Recognition}
Camera-based HAR commonly relies on RGB, skeleton, or multimodal representations. RGB methods capture appearance, scene context, and object interactions, ranging from temporal segment networks~\cite{temporalsegmentnets} and two-stream architectures~\cite{twostreamconvnets} to 3D CNNs~\cite{Feichtenhofer2019} and video transformers~\cite{vivit}. Recent pretrained models such as VideoMAE~\cite{tong2022videomae}, CLIP~\cite{clip}, and DINOv3~\cite{dinov3} provide strong visual representations for image and video understanding. Skeleton-based methods instead operate on body joint coordinates, making them less sensitive to background and appearance changes, though dependent on pose quality. Graph-based models such as ST-GCN~\cite{stgcnpaper}, MS-G3D~\cite{msg3dpaper}, and DG-STGCN~\cite{dgstgcn}, as well as heatmap-based models such as PoseC3D~\cite{Duan2022}, have shown strong performance. 2D-to-3D lifting methods such as MotionAGFormer~\cite{motionagformer} further enable 3D skeleton recognition from monocular video.

Multimodal fusion methods combine complementary cues from RGB and skeleton streams. Simple approaches include feature concatenation and late score fusion~\cite{concatfusion}, while more advanced methods balance modality learning through gradient modulation~\cite{ogmge}, input-quality weighting~\cite{qmf}, modality dropout~\cite{moddrop}, or multi-modality co-learning~\cite{mmcl}. However, robustness under domain shift remains a key challenge for both RGB-based models and fusion methods.

We evaluate these approaches on SAFER-Activities, which is comparable in scale to established HAR benchmarks such as NTU~RGB+D~\cite{Shahroudy2016} (56,880 instances, 40 subjects), PKU-MMD~\cite{Liu2017-nx-pkummd} (21,545 instances, 66 subjects), and Epic-Kitchens-100~\cite{Damen2022RESCALING} (89,977 instances, 37 subjects). It additionally features frame-level annotations, untrimmed multi-camera recordings, and dedicated in-distribution and out-of-distribution evaluation splits. Accordingly, our benchmark spans models ranging from a lightweight 1D~CNN baseline to state-of-the-art skeleton architectures, paired with widely adopted pretrained RGB backbones used as frozen feature extractors.

\subsection{Human Pose Estimation}

The primary benchmarks for human pose estimation (HPE) include the MS COCO Keypoint Detection \cite{Lin2014} and MPII Human Pose \cite{Andriluka2014} datasets, which contain about 200,000 images (COCO) and 25,000 images (MPII) for pose estimation. The OCHuman dataset \cite{ochumanpaper} aims to tackle the challenge of occlusion, with a collection of 5,081 images featuring heavily occluded humans. Likewise, the CrowdPose dataset \cite{Li2019} comprises about 20,000 images of humans in highly crowded scenes. SAFER-Activities features a specialized HPE data subset targeting people in wheelchairs, designed to benchmark HPE models' effectiveness in estimating their poses from diverse camera angles under occlusion.

\section{Dataset}
\label{sec:dataset}

SAFER-Activities contains a collection of videos with frame-level annotations, tailored for action recognition and smart monitoring of people. It features a separate subset focusing on people in wheelchairs for action recognition and pose estimation. We describe the data collection methods, annotation procedures, and key statistics in this section; supplementary material provides additional details on the annotation tool, full action class definitions and illustrations, the visual feature extraction pipeline, and wheelchair pose estimation benchmarks.

\subsection{Frame-level Labels}

We define frame-level labels as those that precisely mark the start and end of an action amidst a sequence of actions. Frame-level labels allow models to exploit temporal dependencies between actions. Take, for instance, a classification model \(M_1\) tasked with predicting an action \(A_t = M_1(F_{t-i}, \ldots, F_t, \ldots, F_{t+k})\) at a given time \(t\), within a window spanning frames \(F_{t-i}, \ldots, F_t, \ldots, F_{t+k}\), (\(i, k \geq 0\)). Through such labeling and the use of untrimmed videos, not all frames in a given window are required to represent the same action, enabling the model to capture information from adjacent actions. Frame-level labels make it more feasible to train models for online action recognition, predicting current actions as they unfold using the temporal context provided by surrounding frames.

\subsection{Dataset Construction}
\label{sec:data_collection}
\begin{figure}[tb]
  \centering
  \begin{subfigure}{0.24\linewidth}
    \includegraphics[width=\linewidth]{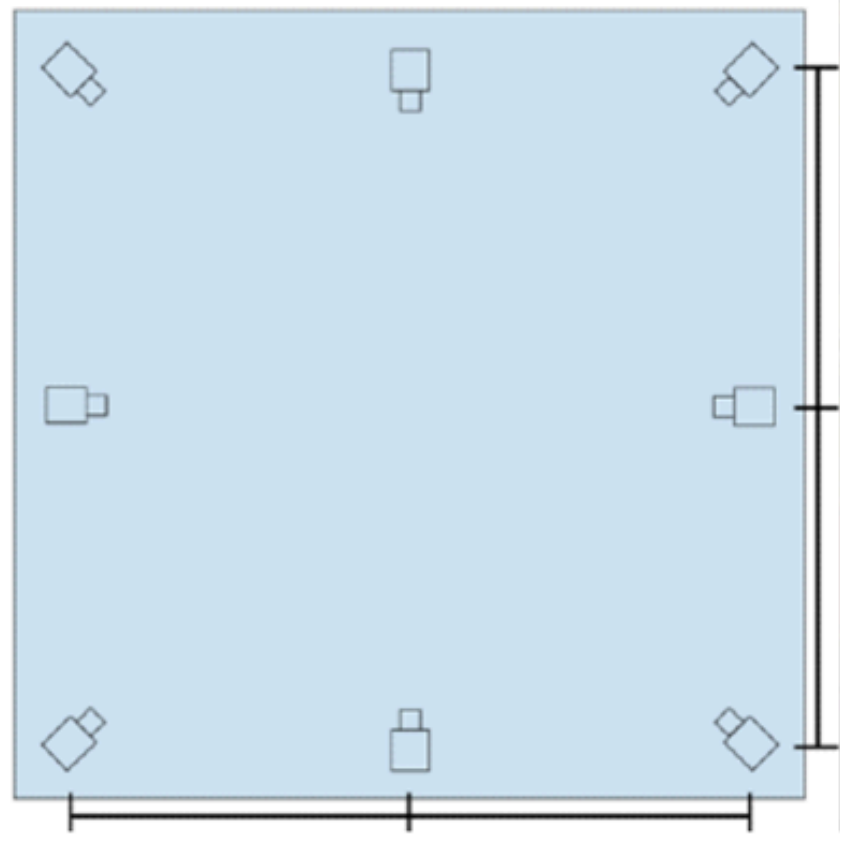}
    \caption{Camera setup}
    \label{fig:dataset-camera-setup}
  \end{subfigure}
  \hfill
  \begin{subfigure}{0.73\linewidth}
    \includegraphics[width=\linewidth]{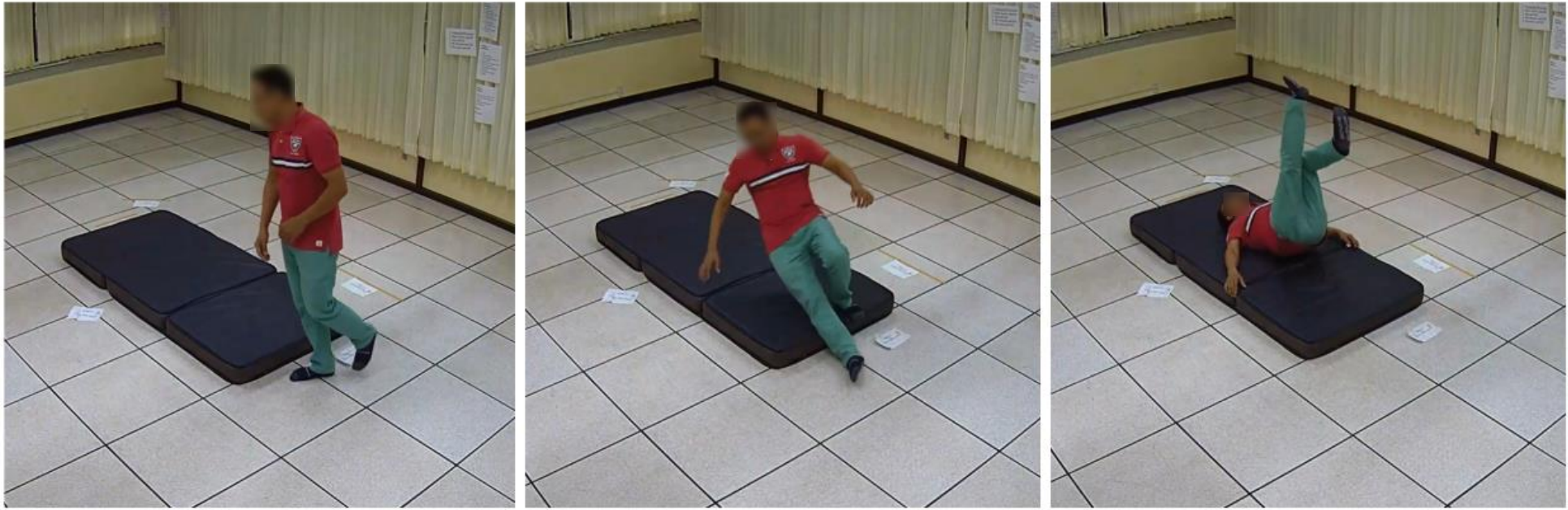}
    \caption{Example fall sequence}
    \label{fig:dataset-action-sequence}
  \end{subfigure}
   \begin{subfigure}{1\linewidth}
    \includegraphics[width=\linewidth]{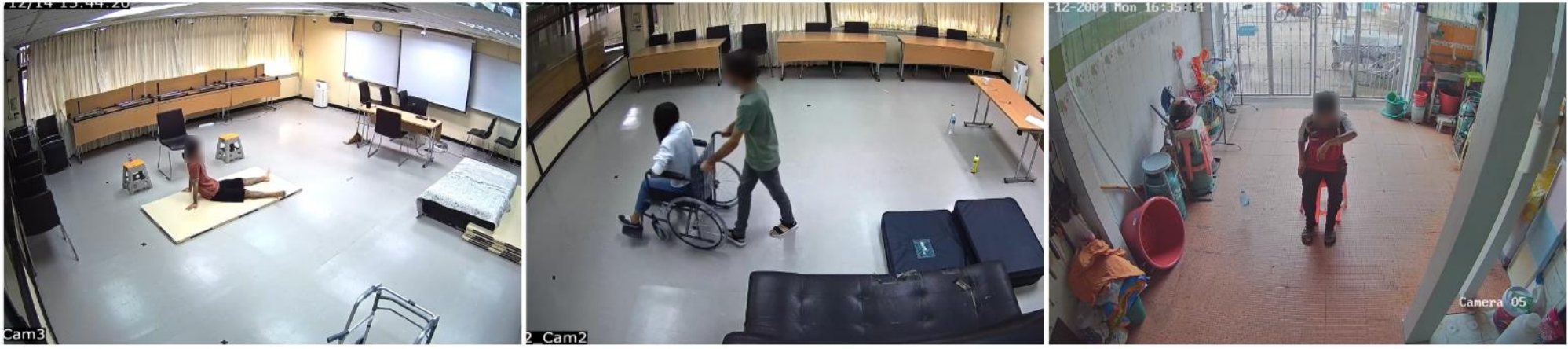}
    \caption{Sample views and actions from in-lab non-wheelchair (left), wheelchair (center), and non-lab (right) subsets}
    \label{fig:action_samples}
  \end{subfigure}
  \caption{Room setup and sample actions}
  \label{fig:dataset-camera-setup-view}
\end{figure}

We performed data collection in four stages, each designed to capture a range of actions, from basic actions such as sitting, pointing, and clapping to critical actions such as falls. The second stage focused on the wheelchair subset, while the first, third, and fourth stages covered activities without wheelchairs. The setup for data collection for the first three stages featured a room with eight fixed high-resolution CCTV cameras aimed toward the center from different directions. We refer to these as the in-lab subsets. Stage 1 data were recorded at 25 fps with a resolution of $2688\times 1520$, whereas for Stages 2, 3, and 4, we increased the resolution to $3072\times2048$. The fourth set (non-lab subset) was exclusively collected in an external home environment with a similar setup but using six cameras instead of eight. Upon entry to the room, participants performed a sequence of predefined activities, guided by audio instructions from a playback device. \Cref{fig:dataset-camera-setup} illustrates the setup of the cameras and room arrangement. \Cref{fig:dataset-camera-setup-view} provides some example actions and views taken from the dataset.

\paragraph{Ethics and Privacy.}
\label{sec:ethics_and_privacy}
This study was approved by the research ethics review committee at the Asian Institute of Technology (Ref. No.: RERC 2022/011). Participants reviewed and signed an informed consent form for their images to be publicly shared for research purposes prior to the data collection. They are free to ask for the removal of their personally identifiable information (face).

\paragraph{Data Annotation.}
To ensure precise action recognition, we developed an annotation tool to mark the start and end of each action in a video. The annotator was provided with a set of predefined labels to cover the expected actions and instructed to decide when each action starts and ends by watching the video closely, going back and forth if necessary to be certain. For example, a ``fall'' usually starts after a ``stand'' or ``unstable'' action when someone begins to fall, but the annotator must check the subsequent few seconds of the video to ensure that the action is a true fall. Once the person is on the ground, the ``fall'' action ends, and the ``lie down'' action begins. Following standard practice in temporally dense activity annotation~\cite{Liu2017-nx-pkummd}, each video was labeled in its entirety by a single annotator to ensure temporal consistency across action boundaries, but independently verified by a second annotator who reviewed the labels and flagged any inconsistencies. Both annotators are computer science M.S.\ graduates working as researchers. To quantify reliability, this second annotator additionally re-annotated approximately 5\% of the recordings from scratch, covering all action classes. Frame-level inter-annotator agreement reached a Cohen's $\kappa$ of 0.89 (0.87 fall/non-fall, 0.87 non-wheelchair, 0.91 wheelchair), with a boundary F1 of 0.86 at a 500~ms tolerance and a median boundary deviation of 145~ms.

\begin{figure}[tb]
  \centering
  \includegraphics[width=\linewidth]{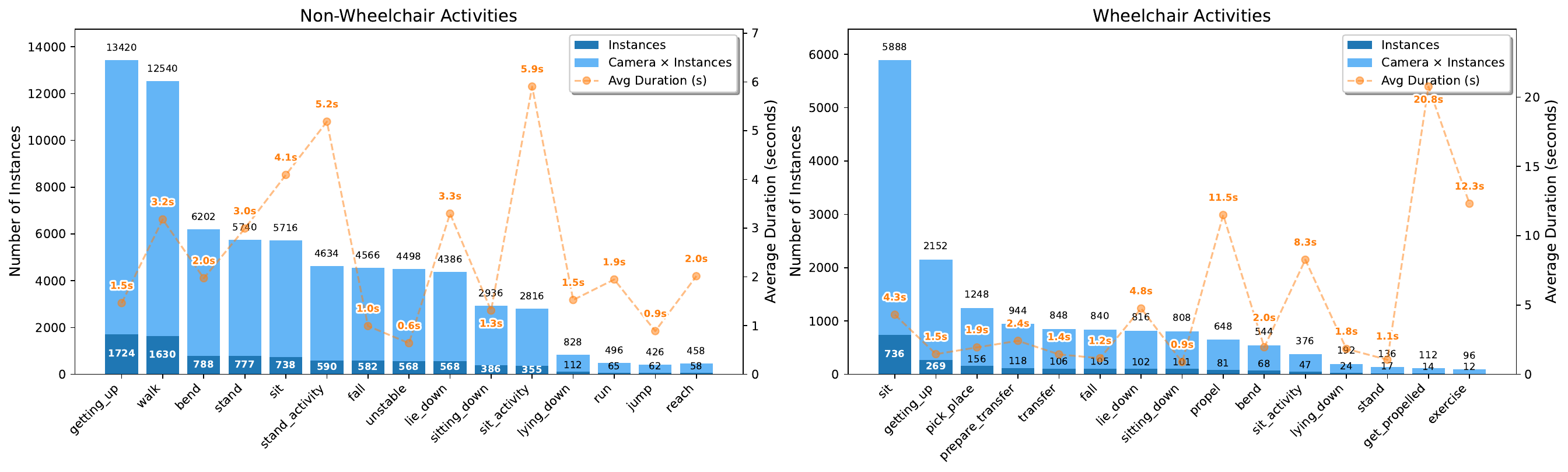}
  \caption{Class statistics for the non-wheelchair (left) and wheelchair (right) subsets.
  }
  \label{fig:action_instances}
\end{figure}

\paragraph{Actions.}
\label{sec:actions}
The final lists of actions we used for model training and analysis are provided in \cref{fig:action_instances}. In our experiments, we generalized common actions to focus the model on more precise fall detection and actions related to physical activity monitoring. For instance, various sitting actions such as clapping, checking the time, making phone calls, waving, and pointing are merged into a single ``sit activity'' macro action. This results in 30 unique actions across both non-wheelchair and wheelchair subsets, down to 15 each from the original 25 and 37, respectively. Descriptions for all the original actions, as well as the illustrations for the generalized macro actions are included in the supplement.

\paragraph{Pose Skeletons and Bounding Boxes.}
\label{sec:pose_skeleton_extraction}
For skeleton-based action recognition, we extract the 2D poses used for training models, following the top-down approach with YOLOv8x~\cite{Jocher_Ultralytics_YOLO_2023} as the detector and the ViTPose-h-multi variant of ViTPose~\cite{xu2022vitpose} as the pose estimator. The pose data are stored following the MMAction2 format \cite{2020mmaction2}. We additionally provide 3D pose sequences obtained by lifting the detected 2D keypoints using a pretrained MotionAGFormer~\cite{motionagformer} network.

\paragraph{Visual Features.}
To complement the skeleton data, we extract visual features from person-centric crops using three frozen backbones: VideoMAE~\cite{tong2022videomae} for spatiotemporal features and CLIP~\cite{clip} and DINOv3~\cite{dinov3} for frame-level features. Crops are obtained by tracking person bounding boxes across frames and resizing to $640\times480$. Features are extracted per-video and temporally aligned with the skeleton sequences to enable direct comparison and multimodal fusion.

\begin{figure}[tb]
  \centering
  \includegraphics[width=\linewidth]{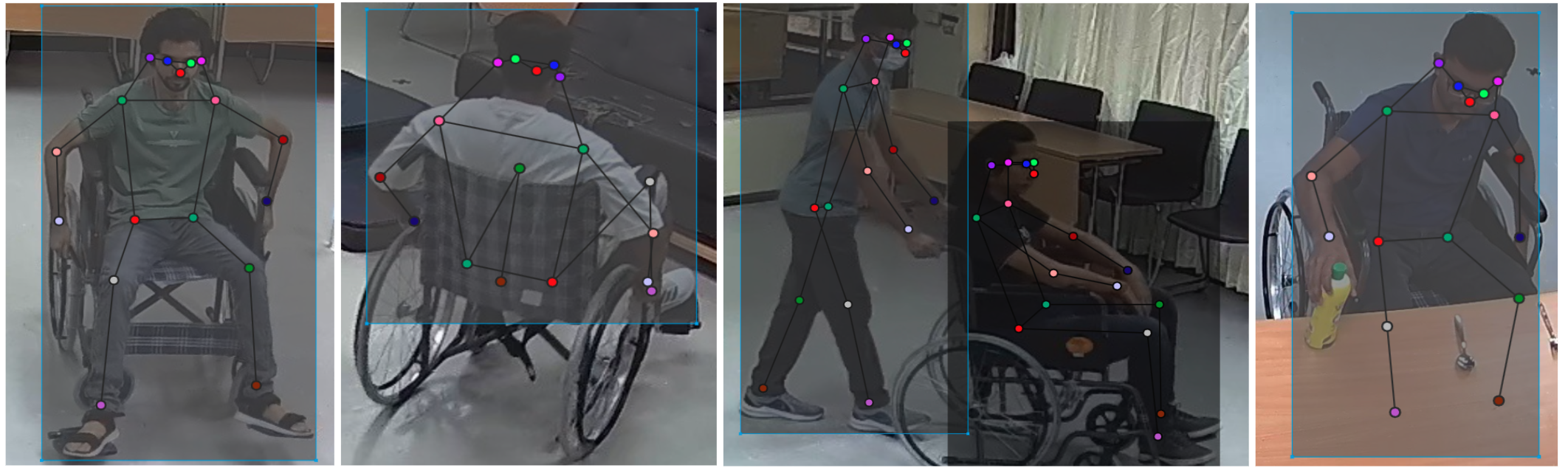}
  \caption{Sample images from the wheelchair keypoints dataset. It contains annotations of people in wheelchairs from various angles and includes highly occluded scenes.
  }
  \label{fig:wheelchairsamples}
\end{figure}

\paragraph{Wheelchair Keypoint Estimation.}
SAFER-Activities includes a complementary pose estimation dataset derived from the wheelchair video data (\cref{fig:wheelchairsamples}). It features 6,846 images containing 8,324 manually-annotated human instances. We used the COCO Annotator tool \cite{cocoannotator} to label person bounding boxes and keypoints. While we have not used this data to fine-tune any models, we do provide the performance of popular HPE methods on this subset. Future work could exploit these annotations to further improve the best keypoint detection models.

\subsection{Statistics}
\label{sec:statistics}
SAFER-Activities comprises over 66 hours of labeled video data, with about 15 hours from the wheelchair subset. The data feature a diverse group of 46 participants. This includes 29 male and 17 female individuals, with an average age of approximately 31 years, ranging from 18 to 58. \Cref{fig:action_instances} shows the total number of action instances and average action duration. ``Camera $\times$ Instances'' refers to the aggregation of multiple views, resulting in multiples of action instances.

\paragraph{Dataset Split}
Following Shahroudy~\etal~\cite{Shahroudy2016}, we split both datasets by subject and camera view. The non-wheelchair dataset (467 videos) was divided into training/testing sets of 371/96 (subject-wise) and 350/117 (view-wise). The wheelchair dataset (142 videos) was split into training/testing sets of 96/46 (subject-wise) and 104/38 (view-wise). Additionally, 30 non-lab test videos from 5 participants were recorded in a home environment with different lighting conditions, camera viewpoints, and scenes with partial occlusion, none of which appear in the training data. This split serves as an out-of-distribution (OOD) evaluation to assess generalization beyond the controlled lab setting.

\section{Experiments}
\label{sec:experiments}

We evaluate action recognition on SAFER-Activities using 2D and 3D skeleton models, frozen RGB-only models, and fusion approaches. Results are reported on the in-lab non-wheelchair, non-lab (OOD), and wheelchair test sets; subsequent tables label these splits as \emph{In-Lab}, \emph{Non-Lab}, and \emph{Wheelchair}, respectively. We additionally perform cross-dataset evaluation on an external dataset and also analyze the effect of fusion strategies and the temporal context window size.

\subsection{Experimental Setup}
\label{sec:training}

All models operate on the skeleton and visual features described in \cref{sec:pose_skeleton_extraction}. We use the subject-wise splits and report macro-averaged and per-class F1-scores. We additionally report segment-level F1 at IoU thresholds in the supplement.

\paragraph{Input and label assignment.}
Following Duan~\etal~\cite{Duan2022}, a sliding window extracts 48-frame sequences from each video. During training, a sub-clip is randomly sampled every 20 frames; at test time, a dense sliding window covers the full video. Labels are assigned by majority vote over five frames centered on the window midpoint. The visual branch of frozen RGB and fusion models uses 16 frames, uniformly subsampled from the same 48-frame window.

\paragraph{Training protocol.}
For skeleton-based recognition, we evaluate a lightweight 1D~CNN (CNN1D) on normalized joint coordinates, ST-GCN++~\cite{duan2022pyskl} and MS-G3D~\cite{msg3dpaper} on both 2D and lifted 3D poses, PoseC3D~\cite{Duan2022} on 2D, and DG-STGCN~\cite{dgstgcn} on 3D. For RGB-only models, each frozen visual backbone is paired with a linear classifier; CLIP~\cite{clip} and DINOv3~\cite{dinov3} features are temporally mean-pooled across frames, similar to Oquab~\etal~\cite{dinov2}, while VideoMAE~\cite{tong2022videomae} directly produces a single clip-level representation. Multimodal fusion combines each visual backbone with the CNN1D skeleton stream via feature concatenation, keeping the skeleton branch fixed for a controlled comparison across fusion variants. The PYSKL framework~\cite{duan2022pyskl} trains all GCN-based and PoseC3D skeleton models with SGD and cosine annealing~\cite{cosineannealing}, following the default PYSKL configurations, while RGB-only and fusion models use AdamW~\cite{adamw} with cosine annealing. Additional details for each model can be found in the training configuration files.

\subsection{Baseline Results}
\label{sec:baseline_results}

\begin{table*}[!tb]
  \centering
  \caption{Per-class F1 scores (\%) on the non-wheelchair test sets. Top: in-lab, bottom: non-lab (OOD). \textbf{Bold} = best per column within each subset; \underline{underline} = second best.}
  \label{tab:nonwheelchair_f1_merged}
  \resizebox{\textwidth}{!}{
  \begin{tabular}{@{}c@{\hspace{4pt}}l*{15}{c}>{\columncolor{gray!15}}c@{}}
    \toprule
    & Model & GU & W & B & S & SA & U & ST & F & LD & SIA & SD & LDN & R & RN & J & \cellcolor{white}Avg \\
    \midrule
    \multicolumn{18}{l}{\textit{In-Lab Test Set}} \\
    \midrule
    \multirow{4}{*}{\rotatebox[origin=c]{90}{\scriptsize 2D}} & CNN1D & 88.7 & 94.1 & 84.5 & 83.8 & 96.7 & 71.8 & 87.5 & 92.3 & 93.4 & 91.3 & 79.5 & 56.9 & 70.3 & 70.9 & 94.9 & 83.8 \\
    & ST-GCN++         & \underline{89.4} & 95.0 & 85.6 & 87.7 & 97.3 & 70.8 & 86.5 & \underline{92.9} & \underline{94.1} & 93.7 & \textbf{85.0} & 62.5 & 79.8 & \underline{87.2} & \underline{96.9} & \underline{86.9} \\
    & PoseC3D          & 88.8 & \underline{95.5} & 86.0 & 86.5 & \textbf{97.9} & 71.7 & 88.4 & \textbf{93.6} & 93.9 & 93.4 & 84.0 & 62.4 & 78.1 & 80.1 & \textbf{98.3} & 86.6 \\
    & MS-G3D           & 89.1 & 95.2 & 86.8 & \textbf{89.0} & \underline{97.8} & 74.1 & 87.6 & 92.7 & \underline{94.1} & \textbf{95.1} & \underline{84.7} & 61.2 & 78.6 & \textbf{88.9} & 93.4 & \textbf{87.2} \\
    \midrule
    \multirow{3}{*}{\rotatebox[origin=c]{90}{\scriptsize 3D}} & ST-GCN++ & 88.8 & 94.0 & 84.3 & 83.0 & 97.0 & 48.6 & 87.3 & 91.8 & \underline{94.1} & 90.6 & 80.6 & \underline{68.9} & 67.7 & 80.4 & 82.9 & 82.7 \\
    & DG-STGCN         & 89.0 & 92.9 & 83.4 & 82.5 & 96.8 & \textbf{74.8} & 86.3 & 92.7 & 88.6 & 89.9 & 80.5 & 66.7 & 43.1 & 80.5 & 82.9 & 82.1 \\
    & MS-G3D           & \textbf{89.8} & 94.9 & 87.3 & 83.1 & 97.4 & \underline{74.5} & 88.7 & 92.4 & 93.8 & 91.5 & 83.4 & \textbf{69.3} & 72.9 & 81.4 & 91.1 & 86.1 \\
    \midrule
    \multirow{3}{*}{\rotatebox[origin=c]{90}{\scriptsize RGB}} & CLIP & 65.3 & 91.5 & 79.2 & 79.7 & 94.6 & 33.9 & 79.6 & 84.9 & 91.1 & 93.5 & 40.2 & 16.6 & 65.2 & 63.9 & 80.3 & 70.6 \\
    & DINOv3           & 69.9 & 92.9 & 81.5 & 78.3 & 94.0 & 35.9 & 78.9 & 87.8 & 91.8 & 90.2 & 47.6 & 19.6 & 72.1 & 61.8 & 79.7 & 72.1 \\
    & VideoMAE         & 80.2 & 94.9 & 81.2 & 79.1 & 95.5 & 61.9 & 84.2 & 89.5 & 91.6 & 92.6 & 62.4 & 51.2 & 78.1 & 69.3 & 93.6 & 80.3 \\
    \midrule
    \multirow{3}{*}{\rotatebox[origin=c]{90}{\scriptsize Fuse}} & CLIP & 86.4 & 94.2 & 87.1 & 86.6 & 97.3 & 71.6 & 89.0 & 90.9 & 93.9 & \underline{94.8} & 79.3 & 57.5 & \underline{83.9} & 66.4 & 96.3 & 85.0 \\
    & DINOv3           & 87.8 & \textbf{95.9} & \textbf{88.4} & 85.7 & 97.7 & 73.0 & \textbf{90.8} & 92.3 & \textbf{94.7} & 93.4 & 82.4 & 59.0 & 81.9 & 65.7 & 95.0 & 85.6 \\
    & VideoMAE         & 88.6 & 95.0 & \underline{88.1} & \underline{87.9} & 97.4 & 70.9 & \underline{89.7} & 92.4 & \underline{94.1} & 94.7 & 81.4 & 56.0 & \textbf{85.9} & 69.7 & 95.1 & 85.8 \\
    \midrule
    \multicolumn{18}{l}{\textit{Non-Lab Test Set (OOD)}} \\
    \midrule
    \multirow{4}{*}{\rotatebox[origin=c]{90}{\scriptsize 2D}} & CNN1D & 89.8 & 91.5 & 72.6 & 70.7 & 78.6 & 41.1 & 72.9 & 71.2 & 92.9 & 73.0 & 83.8 & 65.7 & 31.7 & 83.2 & 88.4 & 73.8 \\
    & ST-GCN++         & \underline{92.1} & 93.0 & \underline{75.7} & \textbf{75.6} & 81.6 & 43.1 & 73.4 & \textbf{80.5} & \textbf{94.7} & 74.6 & 87.5 & 67.3 & 26.6 & 90.6 & 94.8 & 76.7 \\
    & PoseC3D          & 91.1 & \textbf{93.8} & \textbf{76.8} & \underline{75.0} & \textbf{84.6} & \textbf{45.5} & \textbf{77.5} & 79.3 & \underline{94.6} & \textbf{85.6} & \underline{87.6} & 60.5 & \underline{48.1} & \textbf{92.9} & \textbf{96.5} & \textbf{79.3} \\
    & MS-G3D           & 90.9 & \underline{93.7} & 72.2 & 73.7 & \underline{83.9} & 41.1 & \underline{75.1} & 80.1 & 93.2 & \underline{77.9} & \textbf{88.5} & \underline{70.1} & 37.7 & \underline{91.7} & \underline{95.8} & \underline{77.7} \\
    \midrule
    \multirow{3}{*}{\rotatebox[origin=c]{90}{\scriptsize 3D}} & ST-GCN++ & 90.2 & 91.6 & 71.7 & 70.7 & 71.2 & 38.7 & 65.4 & 74.3 & 91.9 & 77.7 & 80.6 & 62.4 & 16.5 & 86.5 & 84.5 & 71.6 \\
    & DG-STGCN         & 90.7 & 90.0 & 67.6 & 71.3 & 74.3 & 38.2 & 67.6 & 78.1 & 92.9 & 75.3 & 81.2 & \textbf{70.2} & 32.0 & 79.2 & 79.9 & 72.6 \\
    & MS-G3D           & \textbf{92.2} & 92.2 & 73.2 & 70.9 & 75.5 & \underline{44.4} & 69.5 & \underline{80.3} & 92.5 & 72.7 & 85.4 & 60.5 & \textbf{51.9} & 85.3 & 83.8 & 75.3 \\
    \midrule
    \multirow{3}{*}{\rotatebox[origin=c]{90}{\scriptsize RGB}} & CLIP & 46.9 & 72.7 & 29.3 & 53.1 & 69.3 & 0.0  & 39.6 & 9.9  & 84.3 & 69.7 & 17.8 & 11.9 & 0.0  & 28.7 & 0.0  & 35.6 \\
    & DINOv3           & 24.0 & 24.5 & 27.0 & 46.6 & 22.4 & 0.0  & 42.1 & 0.0  & 81.7 & 68.5 & 7.6  & 11.0 & 0.0  & 0.2  & 1.2  & 23.8 \\
    & VideoMAE         & 69.7 & 79.2 & 26.0 & 55.9 & 68.8 & 37.4 & 56.0 & 49.0 & 84.7 & 69.3 & 39.5 & 17.1 & 22.8 & 25.4 & 30.4 & 48.7 \\
    \midrule
    \multirow{3}{*}{\rotatebox[origin=c]{90}{\scriptsize Fuse}} & CLIP & 81.7 & 81.7 & 51.7 & 58.9 & 76.5 & 15.5 & 59.2 & 58.8 & 89.9 & 72.6 & 59.8 & 54.7 & 3.3  & 19.8 & 0.0  & 52.3 \\
    & DINOv3           & 70.4 & 67.5 & 43.3 & 50.3 & 11.8 & 13.3 & 59.9 & 47.5 & 87.1 & 3.2  & 14.9 & 23.9 & 0.0  & 0.0  & 0.0  & 32.9 \\
    & VideoMAE         & 80.0 & 82.5 & 50.3 & 56.4 & 70.6 & 25.1 & 54.9 & 66.2 & 87.1 & 66.8 & 59.9 & 35.9 & 0.5  & 57.9 & 23.0 & 54.5 \\
    \bottomrule
  \end{tabular}
  }
  \caption*{\scriptsize 2D/3D = skeleton input (3D via MotionAGFormer~\cite{motionagformer} lifting), RGB = frozen pretrained features, Fuse = visual backbone + CNN1D feature concatenation; GU = Getting Up, W = Walk, B = Bend, S = Sit, SA = Standing Activity, U = Unstable, ST = Stand, F = Fall, LD = Lie Down, SIA = Sitting Activity, SD = Sitting Down, LDN = Lying Down, R = Reach, RN = Run, J = Jump.}
\end{table*}

\begin{table*}[!tbp]
  \centering
  \caption{Per-class F1 scores (\%) on the wheelchair test set. \textbf{Bold} = best per column; \underline{underline} = second best.}
  \label{tab:wheelchair_f1}
  \resizebox{\textwidth}{!}{
  \begin{tabular}{@{}c@{\hspace{4pt}}l*{15}{c}>{\columncolor{gray!15}}c@{}}
    \toprule
    & Model & S & GU & PP & PT & TR & F & LD & SD & PR & B & SIA & LDN & ST & GP & E & \cellcolor{white}Avg \\
    \midrule
    \multirow{4}{*}{\rotatebox[origin=c]{90}{\scriptsize 2D}} & CNN1D & 86.1 & 77.3 & 56.5 & 65.6 & 55.2 & 76.7 & 89.1 & 70.9 & 86.2 & 67.5 & 70.3 & 41.9 & 29.1 & 67.0 & 83.1 & 68.2 \\
    & ST-GCN++        & 89.8 & 80.3 & 66.9 & 64.1 & 61.5 & 81.2 & 91.2 & 71.2 & 93.7 & 69.3 & 71.3 & 49.7 & 16.1 & 87.9 & 84.9 & 71.9 \\
    & PoseC3D         & 90.3 & 81.2 & 69.9 & \underline{70.2} & 62.4 & \textbf{87.0} & 90.1 & 71.3 & 94.6 & 74.8 & \underline{81.3} & 61.2 & 20.4 & 90.1 & \underline{92.8} & 75.8 \\
    & MS-G3D          & 90.3 & \textbf{82.5} & \underline{77.9} & 69.6 & 63.4 & \underline{86.7} & 91.7 & \textbf{74.9} & 94.4 & 74.6 & 80.0 & \textbf{67.1} & 29.8 & 88.4 & 90.7 & \underline{77.5} \\
    \midrule
    \multirow{3}{*}{\rotatebox[origin=c]{90}{\scriptsize 3D}} & ST-GCN++ & 87.7 & 80.1 & 70.1 & 68.8 & 63.9 & 81.8 & 92.5 & 70.5 & 87.1 & 73.7 & 51.3 & 55.3 & 22.8 & 92.3 & 87.6 & 72.4 \\
    & DG-STGCN        & 87.9 & 80.2 & 68.9 & 69.6 & 62.8 & 80.5 & 92.1 & 71.8 & 88.9 & 77.5 & 70.1 & 57.6 & 31.8 & 91.9 & 84.8 & 74.4 \\
    & MS-G3D          & 87.8 & 80.1 & 68.6 & 68.3 & 63.9 & 81.9 & 73.4 & \underline{74.8} & 87.8 & 74.7 & 71.8 & \underline{64.0} & 32.3 & 92.1 & 88.3 & 74.0 \\
    \midrule
    \multirow{3}{*}{\rotatebox[origin=c]{90}{\scriptsize RGB}} & CLIP & 86.9 & 48.9 & 65.3 & 41.2 & 30.1 & 41.3 & 89.9 & 44.4 & 86.8 & 62.4 & 66.9 & 29.7 & 25.4 & 95.6 & 88.2 & 60.2 \\
    & DINOv3          & 87.7 & 51.5 & 58.2 & 53.6 & 55.8 & 59.5 & 91.0 & 44.3 & 90.1 & 64.0 & 61.4 & 28.6 & 36.5 & 95.9 & 91.1 & 64.6 \\
    & VideoMAE        & 89.6 & 68.4 & 77.6 & 55.7 & 60.6 & 78.0 & 90.8 & 53.7 & \textbf{96.1} & 74.4 & 78.5 & 35.4 & 21.8 & \textbf{99.4} & 91.4 & 71.4 \\
    \midrule
    \multirow{3}{*}{\rotatebox[origin=c]{90}{\scriptsize Fuse}} & CLIP & 89.4 & 79.6 & 69.1 & 63.9 & 62.2 & 80.3 & \underline{94.1} & 72.3 & 88.9 & 73.2 & 76.7 & 55.2 & 37.3 & 93.9 & 90.4 & 75.1 \\
    & DINOv3           & \underline{91.4} & 80.0 & 72.5 & \textbf{72.4} & \textbf{67.7} & 80.4 & \underline{94.1} & 69.6 & 92.6 & \underline{82.7} & 76.0 & 58.4 & \textbf{50.7} & 94.5 & \textbf{95.4} & \textbf{78.6} \\
    & VideoMAE         & \textbf{92.2} & \underline{82.1} & \textbf{80.5} & 69.5 & \underline{66.6} & 80.3 & \textbf{94.9} & 74.0 & \underline{95.9} & \textbf{83.9} & \textbf{81.4} & 50.3 & \underline{38.5} & \underline{99.3} & 90.4 & \textbf{78.6} \\
    \bottomrule
  \end{tabular}
  }
  \caption*{\scriptsize 2D/3D = skeleton input (3D via MotionAGFormer~\cite{motionagformer} lifting), RGB = frozen pretrained features, Fuse = visual backbone + CNN1D feature concatenation; S = Sit, GU = Getting Up, PP = Pick/Place, PT = Prepare Transfer, TR = Transfer, F = Fall, LD = Lie Down, SD = Sitting Down, PR = Propel, B = Bend, SIA = Sitting Activity, LDN = Lying Down, ST = Stand, GP = Get Propelled, E = Exercise.}
\end{table*}

\Cref{tab:nonwheelchair_f1_merged,tab:wheelchair_f1} report per-class F1-scores across all modalities and test sets. On the in-lab splits, 2D skeleton models achieve the highest overall scores, with MS-G3D reaching 87.2\% on the non-wheelchair subset and 77.5\% on the wheelchair subset. 3D pose models are competitive but slightly behind their 2D counterparts. The frozen RGB-only models lag substantially: VideoMAE is the strongest at 80.3\% and 71.4\%, while CLIP and DINOv3 trail further. Fusion is competitive with skeleton models in-lab (up to 85.8\% non-wheelchair) and achieves the best wheelchair results, with DINOv3 and VideoMAE fusion both reaching 78.6\%. Notably, all modalities detect falls reliably in controlled settings.

The non-lab results reveal a stark modality gap. Frozen RGB features collapse under domain shift: DINOv3 drops from 72.1\% to 23.8\%, CLIP from 70.6\% to 35.6\%, and VideoMAE from 80.3\% to 48.7\%. Skeleton models prove far more robust, with the best 2D score declining from 87.2\% to 79.3\%. Feature-concatenation fusion inherits the RGB weakness and falls below skeleton-only performance. The disparity is most severe for fall detection: RGB models nearly fail entirely (DINOv3 0.0\% F1, CLIP 9.9\%), whereas skeleton models maintain 70--80\% F1.

Several classes remain challenging across all models. ``Unstable'' is consistently the hardest (best in-lab: 74.8\%, best OOD: 45.5\%), likely due to its short average duration and motion patterns that overlap with other actions. ``Lying Down'' and ``Reach'' also degrade sharply under domain shift, the latter possibly confused with ``Bend'' across viewpoints. The gap between the lightweight CNN1D and state-of-the-art models widens in these harder settings, indicating that more expressive architectures better capture the fine-grained temporal and spatial cues needed for robust recognition. These findings point to three open challenges for future work: improving OOD robustness, fusion strategies that do not inherit the visual domain gap, and disambiguating visually similar actions.

\subsection{Additional Experiments}
\label{sec:additional_experiments}
\begin{table}[tb]
  \begin{minipage}[t]{0.54\textwidth}
    \centering
    \caption{Macro F1 (\%) for fusion strategies pairing each visual backbone with CNN1D skeleton features. Dropout = ModDrop~\cite{moddrop}; MMCL~\cite{mmcl} uses skeleton only at inference. \textbf{Bold} = best, \underline{underline} = second best.}
    \label{tab:fusion_analysis}
    \scriptsize
    \begin{tabular*}{\linewidth}{@{\extracolsep{\fill}}c@{\hspace{4pt}}lccc@{}}
      \toprule
      & Backbone & In-Lab & Non-Lab & Wheelchair \\
      \midrule
      & CNN1D (skel.\ only) & 83.8 & \textbf{73.8} & 68.2 \\
      \specialrule{\lightrulewidth}{4pt}{4pt}
      \multirow{3}{*}{\rotatebox[origin=c]{90}{\tiny Concat}} & DINOv3 & 85.6 & 32.9 & \textbf{78.6} \\
      & CLIP    & 85.0 & 52.3 & 75.1 \\
      & VideoMAE & 85.8 & 54.5 & \textbf{78.6} \\
      \specialrule{\lightrulewidth}{4pt}{4pt}
      \multirow{3}{*}{\rotatebox[origin=c]{90}{\tiny Dropout}} & DINOv3 & 85.4 & 59.6 & 75.4 \\
      & CLIP    & 85.4 & 61.9 & 74.3 \\
      & VideoMAE & \textbf{86.3} & 67.3 & 76.3 \\
      \specialrule{\lightrulewidth}{4pt}{4pt}
      \multirow{3}{*}{\rotatebox[origin=c]{90}{\tiny QMF}} & DINOv3 & 85.6 & 32.5 & 76.6 \\
      & CLIP    & 85.5 & 52.6 & 75.2 \\
      & VideoMAE & \underline{86.2} & 52.4 & 78.1 \\
      \specialrule{\lightrulewidth}{4pt}{4pt}
      \multirow{3}{*}{\rotatebox[origin=c]{90}{\tiny OGMGE}} & DINOv3 & 85.5 & 29.0 & 78.4 \\
      & CLIP    & 83.4 & 45.3 & 75.9 \\
      & VideoMAE & 86.0 & 55.2 & \underline{78.5} \\
      \specialrule{\lightrulewidth}{4pt}{4pt}
      \multirow{3}{*}{\rotatebox[origin=c]{90}{\tiny MMCL}} & DINOv3 & 82.5 & 66.6 & 68.9 \\
      & CLIP    & 82.2 & \underline{68.0} & 69.7 \\
      & VideoMAE & 83.4 & 66.5 & 68.7 \\
      \specialrule{\lightrulewidth}{4pt}{0pt}
      \bottomrule
    \end{tabular*}
  \end{minipage}
  \hfill
  \begin{minipage}[t]{0.42\textwidth}
    \centering
    \caption{Temporal stride (TS) effect on 2D skeleton models (macro F1, \%). \textbf{Bold} = best per column.}
    \label{tab:temporal_stride}
    \scriptsize
    \begin{tabular*}{\linewidth}{@{\extracolsep{\fill}}lcccc@{}}
      \toprule
      & \multicolumn{2}{c}{In-Lab} & \multicolumn{2}{c}{Wheelchair} \\
      \cmidrule(lr){2-3} \cmidrule(l){4-5}
      Model & TS\,1 & TS\,3 & TS\,1 & TS\,3 \\
      \midrule
      CNN1D & 83.8 & 84.3 & 68.2 & 69.3 \\
      ST-GCN++ & 86.9 & 85.7 & 71.9 & 75.6 \\
      PoseC3D & 86.6 & 86.7 & 75.8 & 78.3 \\
      MS-G3D & \textbf{87.2} & \textbf{87.1} & \textbf{77.5} & \textbf{78.5} \\
      \bottomrule
    \end{tabular*}
    \par\bigskip
    \caption{Cross-dataset fall detection on ImViA~\cite{imviadataset} without fine-tuning. \textbf{Bold} = best per column.}
    \label{tab:fall_accuracy_on_existing_datasets}
    \scriptsize
    \begin{tabular*}{\linewidth}{@{\extracolsep{\fill}}c@{\hspace{4pt}}lccc@{}}
      \toprule
      & Model & Precision & Recall & F1 \\
      \midrule
      \multirow{2}{*}{\rotatebox[origin=c]{90}{\scriptsize 2D}} & CNN1D & 92.0 & 94.1 & 93.0 \\
      & PoseC3D & 98.9 & 94.8 & \textbf{96.8} \\
      \midrule
      \multirow{3}{*}{\rotatebox[origin=c]{90}{\scriptsize RGB}} & CLIP & undef. & 0.0 & 0.0 \\
      & DINOv3 & undef. & 0.0 & 0.0 \\
      & VideoMAE & 86.4 & 38.4 & 53.1 \\
      \midrule
      \multirow{3}{*}{\rotatebox[origin=c]{90}{\scriptsize Fuse}} & CLIP & \textbf{100.0} & 66.7 & 80.0 \\
      & DINOv3 & \textbf{100.0} & 64.6 & 78.5 \\
      & VideoMAE & 96.9 & \textbf{96.0} & 96.4 \\
      \bottomrule
    \end{tabular*}
  \end{minipage}
\end{table}

\paragraph{Fusion strategies.}
\Cref{sec:baseline_results} showed that feature concatenation inherits the frozen RGB domain-shift weakness. \Cref{tab:fusion_analysis} compares alternative fusion strategies introduced to mitigate issues in modality fusion across all three visual backbones. ModDrop~\cite{moddrop} is the most effective, substantially recovering non-lab performance: VideoMAE improves from 54.5\% to 67.3\%, DINOv3 from 32.9\% to 59.6\%, and CLIP from 52.3\% to 61.9\%. QMF~\cite{qmf} and OGM-GE~\cite{ogmge} do not consistently improve over concatenation. MMCL~\cite{mmcl}, which discards RGB at inference and uses only skeleton, does not improve non-lab performance over the CNN1D baseline but notably improves wheelchair recognition (e.g., 69.7\% vs.\ 68.2\% for CNN1D alone). Despite these strategies, no fusion method surpasses the skeleton-only CNN1D on the non-lab set (73.8\%), indicating that better fusion strategies are needed to leverage the strengths of each modality under domain shift.

\paragraph{Temporal context window.}
\Cref{tab:temporal_stride} compares temporal stride 1 (48 consecutive frames) and stride 3 (sampling every 3rd frame, covering 144 frames) for the 2D skeleton models. In-lab non-wheelchair performance is nearly identical across strides, but wheelchair recognition improves consistently with a longer context.

\paragraph{Cross-dataset fall detection.}
To evaluate cross-dataset generalization, we test a subset of models trained on SAFER-Activities on the ImViA dataset~\cite{imviadataset}, using the 130 videos (99 falls) that provide frame-level fall annotations. Since ImViA contains only fall annotations, we evaluate fall detection exclusively. Poses and RGB features are extracted using the same pipeline described in \cref{sec:pose_skeleton_extraction}. To assess accuracy, we cluster consecutive fall predictions into temporal events and match them against ground-truth clusters; a tolerance window of 15 frames accounts for labeling differences across datasets. \Cref{tab:fall_accuracy_on_existing_datasets} shows that skeleton models generalize well: PoseC3D achieves 96.8\% F1 and CNN1D reaches 93.0\%. Frozen RGB-only models confirm the domain-shift vulnerability: CLIP and DINOv3 predict no fall events at all, so their precision is undefined and their recall and F1 are 0; VideoMAE, by contrast, suffers from very low recall (38.4\%). Fusion of VideoMAE with CNN1D recovers this gap, nearly matching PoseC3D at 96.4\% F1, whereas CLIP and DINOv3 fusion achieve high precision but limited recall.

\subsection{Qualitative Analysis}

\begin{figure}[!t]
  \centering
  \includegraphics[width=\linewidth]{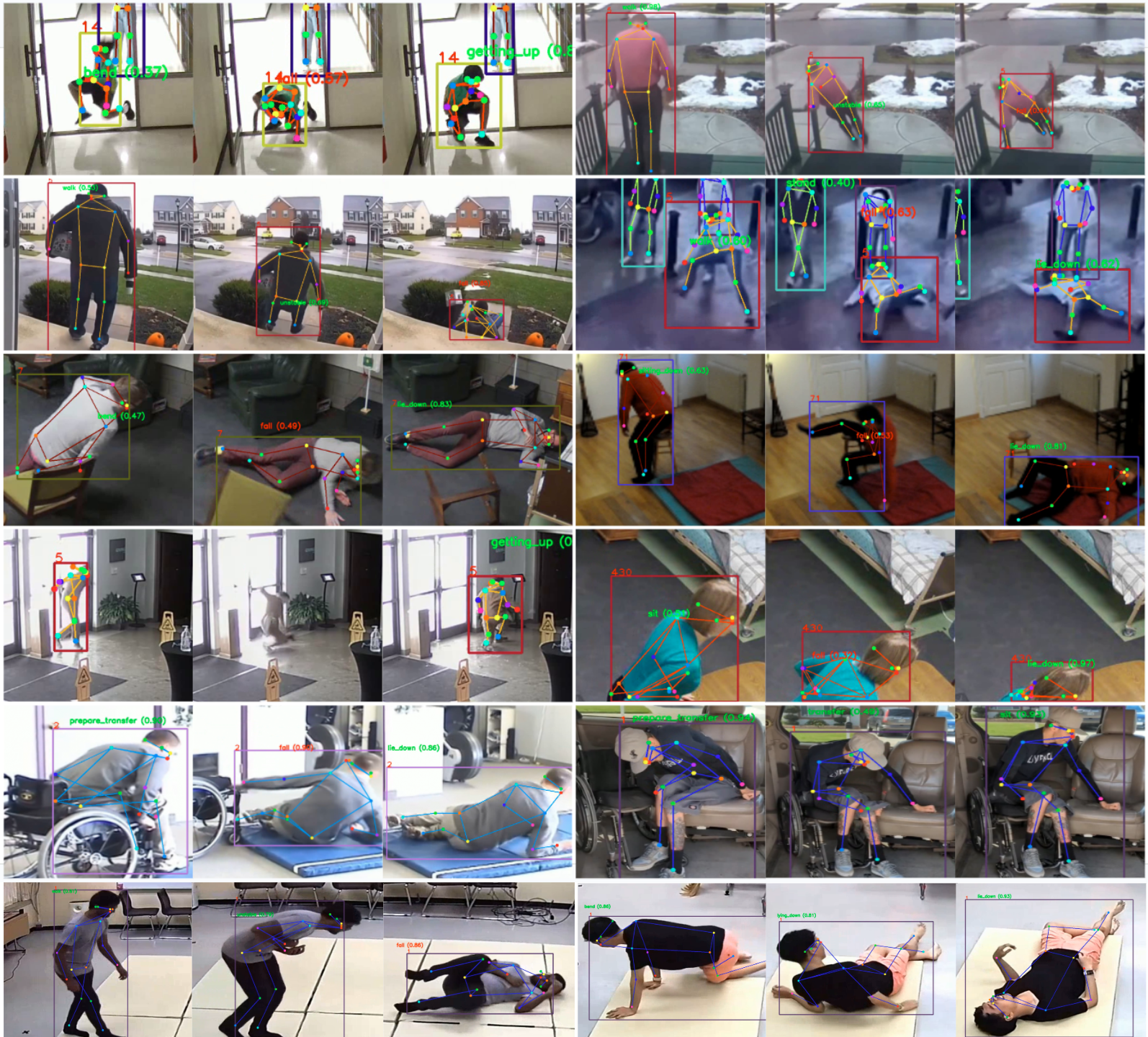}
    \caption{CNN1D predictions on external and in-lab data. Rows 1--2: real falls, row 1 right and row 2 from \cite{failarmy}. Row 3: realistic fall datasets. Row 4: failures from missed detections (left) and severe occlusion (right). Row 5: wheelchair actions from \cite{regor09} (fall, left) and \cite{livetoroll} (transfer, right). Row 6 (in-lab): fall (left) versus slowly lying down (right).}
  \label{fig:qualitativeresults}
\end{figure}

\Cref{fig:qualitativeresults} shows CNN1D predictions on diverse data, including external real-world falls and wheelchair user actions. On external data (rows 1--5), skeleton-based models trained on SAFER-Activities perform well outside the training distribution, with failures primarily due to poor-quality poses under severe occlusion or missed human detections. Notably, the frame-level temporal annotations in SAFER-Activities enable even a lightweight model such as CNN1D to distinguish falls from visually similar actions like slowly lying down on a surface (row 6). This distinction, raised as a key concern for fall detection systems, relies on the abrupt temporal dynamics rather than static pose similarity of actions.

\section{Conclusion}

We presented SAFER-Activities, a large-scale dataset with frame-level labels for online action recognition, especially fall detection, in untrimmed videos. The dataset includes a dedicated wheelchair subset and is released with precomputed pose skeletons and visual features, providing a comprehensive resource for smart healthcare monitoring research. Our evaluation spans 2D and 3D skeleton models, frozen RGB backbones, and multimodal fusion. Skeleton-based models prove the most robust under domain shift, generalizing well to an out-of-distribution home environment and an external fall dataset. Fusing frozen RGB features with the baseline CNN1D improves in-domain recognition, most clearly on the wheelchair subset, but degrades out of distribution; the tested fusion strategies reduce but do not fully close this gap. The dense temporal annotations enable even lightweight models to distinguish visually similar actions such as falls and lying down based on temporal dynamics. We believe that SAFER-Activities, with its multi-modality benchmarks and out-of-distribution evaluation, will stimulate research on robust action recognition for safety-critical healthcare applications.

\paragraph{Limitations and Future Work.}
For safety and ethics reasons, fall simulations were performed by adult actors rather than elderly participants or regular wheelchair users. This design enabled controlled, repeatable capture of diverse fall and fall-like motions, while prospective validation with the intended populations remains an important next step. SAFER-Activities is therefore best viewed as a pre-deployment benchmark for fall dynamics, fall-like routine activities, and wheelchair-use scenarios rather than a clinical validation study. The wheelchair subset focuses on in-lab recordings; without a public external wheelchair activity dataset, our current external evidence for wheelchair-use scenarios is qualitative (\cref{fig:qualitativeresults}). Future expansions with elderly participants, naturalistic falls, and larger non-lab wheelchair-use recordings would further strengthen ecological validity.

Our findings suggest several directions for model development. First, our synchronized recordings enable robust multi-view architectures~\cite{multitsf}. Second, because cross-dataset evaluations reveal that frozen RGB features struggle with appearance shifts, future fusion strategies must adaptively leverage visual context when it is reliable, while defaulting to robust pose dynamics under domain shift. Finally, sharp performance drops on visually ambiguous, safety-critical actions like falls motivate methods that combine pose dynamics with complementary depth or inertial cues to disambiguate difficult cases~\cite{utdmhad}.

\subsubsection{\ackname}
We thank all the volunteers who participated in the data collection. This work was supported by Thailand's office of the National Broadcasting and Telecommunications Commission and the Broadcasting and Telecommunications Research and Development Fund for Public Interest under Grant A64-1-(2)-006.

\bibliographystyle{splncs04}
\bibliography{main}

\clearpage
\appendix

\paragraph{Data and Code Availability.}
SAFER-Activities, together with the benchmark code, precomputed pose skeletons and visual features, and trained model weights, is publicly available through the project page\footnote{\url{https://safer-activities.github.io/}} and on Hugging Face\footnote{\url{https://huggingface.co/datasets/SAFER-Activities/SAFER-Activities}}. Access is granted upon request, in which users state their intended use.

\section{Additional Experimental Details}
\label{sec:supp_experimental}

Algorithm~\ref{preprocessing_algorithm} illustrates the preprocessing steps to generate sub-clips for training and testing, as detailed in Section~4.1. This example uses a larger window of 144 frames, reduced to 48 frames with a stride of 3. For smaller windows, $w=48$ and $t=1$. For training, one random sub-clip is selected every 20 frames, whereas all sub-clips are used for testing. The evaluation process is similar to running inference on the original videos, but with pre-extracted poses and RGB features.

\begin{algorithm}[h!]
\caption{Preprocessing for Sub-clip Extraction}
\begin{algorithmic}[1]
\State \textbf{Input}: A list of annotations for full-length videos, $V$.
\State \textbf{Output}: A list of extracted sub-clips, $C$.
\State \textbf{Parameters}:
\State \quad $s = 48$ \Comment{The standard input size for models}
\State \quad $w = 144$ \Comment{The number of frames in each sliding window}
\State \quad $t = 3$ \Comment{Stride applied when reducing frame count to $s$}
\State \quad $e_{\text{max}} = 5$ \Comment{Maximum allowed erroneous frames with missing keypoints}

\State $C \gets []$ \Comment{Initialize $C$ as an empty list to store valid clips.}

\For{\textbf{each} video annotation $v$ in $V$}
    \For{$i \gets 0$ \textbf{to} $\text{len}(v) - w$}
        \State $clip \gets v[i : i+w]$ \Comment{Extract a sub-clip of $w$ frames starting at $i$}
        \State $clip' \gets clip[::t]$ \Comment{Reduce the sub-clip to $s$ by selecting every $t^{th}$ frame}
        \State $e \gets \text{count\_errors}(clip')$ \Comment{Calculate frames with missing keypoints}
        \If{$e < e_{\text{max}}$}
            \State $label \gets \text{majority\_vote}(clip'[s/2-2 : s/2+3])$
            \State $C.\text{append}((clip', label))$ \Comment{Append the labeled sub-clip to $C$}
        \EndIf
    \EndFor
\EndFor
\State \Return $C$
\end{algorithmic}
\label{preprocessing_algorithm}
\end{algorithm}

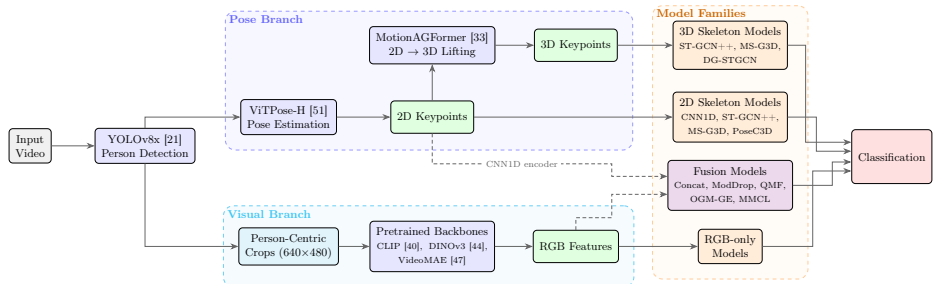
\begin{figure*}[!ht]
\centering
\resizebox{\textwidth}{!}{%
\begin{tikzpicture}[
  >=Stealth,
  every node/.style={font=\small},
  proc/.style={draw, rounded corners=3pt, fill=#1, minimum height=0.9cm, align=center, inner sep=5pt, line width=0.4pt},
  proc/.default={blue!8},
  grplabel/.style={font=\small\bfseries, anchor=north west},
  arr/.style={->, thick, draw=black!60},
  darr/.style={->, thick, draw=black!60, densely dashed},
]

\node[proc=gray!12] (video) at (0, 0) {Input\\Video};

\node[proc=blue!10] (yolo) at (3.2, 0) {YOLOv8x~\cite{Jocher_Ultralytics_YOLO_2023}\\Person Detection};

\node[proc=blue!10] (vitpose) at (7.2, 0.8) {ViTPose-H~\cite{xu2022vitpose}\\Pose Estimation};
\node[proc=cyan!10] (crop) at (7.2, -2.8) {Person-Centric\\Crops ($640{\times}480$)};

\node[proc=blue!10] (lift) at (11.2, 2.8) {MotionAGFormer~\cite{motionagformer}\\2D $\rightarrow$ 3D Lifting};
\node[proc=green!12] (kp2d) at (11.2, 0.8) {2D Keypoints};
\node[proc=blue!10] (backbones) at (11.2, -2.8) {Pretrained Backbones\\{\scriptsize CLIP~\cite{clip},\; DINOv3~\cite{dinov3},}\\{\scriptsize VideoMAE~\cite{tong2022videomae}}};

\node[proc=green!12] (kp3d) at (15.2, 2.8) {3D Keypoints};
\node[proc=green!12] (rgbfeat) at (15.2, -2.8) {RGB Features};

\node[proc=orange!15] (skel3d) at (19.5, 2.8) {3D Skeleton Models\\{\scriptsize ST-GCN++, MS-G3D,}\\{\scriptsize DG-STGCN}};
\node[proc=orange!15] (skel2d) at (19.5, 0.8) {2D Skeleton Models\\{\scriptsize CNN1D, ST-GCN++,}\\{\scriptsize MS-G3D, PoseC3D}};
\node[proc=violet!15] (fusion) at (19.5, -1.1) {Fusion Models\\{\scriptsize Concat, ModDrop, QMF,}\\{\scriptsize OGM-GE, MMCL}};
\node[proc=orange!15] (rgbmdl) at (19.5, -2.8) {RGB-only\\Models};

\node[proc=red!12, minimum width=2cm, minimum height=1.5cm] (output) at (24, -0.3) {Classification};

\draw[arr] (video) -- (yolo);
\draw[arr] (yolo) |- (vitpose);
\draw[arr] (yolo) |- (crop);
\draw[arr] (vitpose) -- (kp2d);
\draw[arr] (crop) -- (backbones);
\draw[arr] (backbones) -- (rgbfeat);

\draw[arr] (kp2d) -- (skel2d);
\draw[arr] (kp2d.north) -- (lift.south);
\draw[arr] (lift) -- (kp3d);
\draw[arr] (kp3d) -- (skel3d);

\draw[arr] (rgbfeat) -- (rgbmdl);

\draw[darr] (kp2d.south) -- (11.2, -0.5) -- (16.2, -0.5)
            -- (16.2, -0.85) -- ([yshift=0.25cm]fusion.west);
\draw[darr] (rgbfeat.north) -- (15.2, -2.0) -- (16.2, -2.0)
            -- (16.2, -1.35) -- ([yshift=-0.25cm]fusion.west);

\node[font=\scriptsize, text=black!60, fill=white, inner sep=2pt] at (13.7, -0.5) {CNN1D encoder};

\draw[arr] (skel3d.east) -- ++(0.5,0) |- ([yshift=0.4cm]output.west);
\draw[arr] (skel2d.east) -- ++(0.8,0) |- ([yshift=0.15cm]output.west);
\draw[arr] (fusion.east) -- ++(1.1,0) |- ([yshift=-0.15cm]output.west);
\draw[arr] (rgbmdl.east) -- ++(1.4,0) |- ([yshift=-0.4cm]output.west);

\begin{scope}[on background layer]
  \node[draw=blue!40, dashed, rounded corners=8pt, fill=blue!3, inner sep=12pt,
        fit=(vitpose)(kp2d)(lift)(kp3d), label={[grplabel, text=blue!60]north west:Pose Branch}] {};
  \node[draw=cyan!50, dashed, rounded corners=8pt, fill=cyan!3, inner sep=12pt,
        fit=(crop)(backbones)(rgbfeat), label={[grplabel, text=cyan!60]north west:Visual Branch}] {};
  \node[draw=orange!40, dashed, rounded corners=8pt, fill=orange!3, inner sep=12pt,
        fit=(skel2d)(skel3d)(rgbmdl)(fusion), label={[grplabel, text=orange!80!black]north west:Model Families}] {};
\end{scope}

\end{tikzpicture}%
}
\caption{Overview of the SAFER-Activities evaluation pipeline. Person skeletons and person-centric crops are extracted using YOLOv8x~\cite{Jocher_Ultralytics_YOLO_2023} bounding boxes and processed through the pose and visual branches.}
\label{fig:pipeline}
\end{figure*}

Figure~\ref{fig:pipeline} illustrates the full evaluation pipeline described in Sections~3 and~4, showing the pose and visual branches, the four model families, and how fusion models combine skeleton and visual streams.

\section{Wheelchair Keypoints Dataset}
\label{sec:supp_wheelchair_hpe}

We evaluated various popular HPE methods on the wheelchair keypoints dataset with COCO-pretrained models from MMPose~\cite{mmpose2020}. The pose estimation accuracy is shown in Table~\ref{tab:wheelchair_model_performance}, using the Percentage of Correct Keypoints (PCK)~\cite{pckposepaper} and Object Keypoint Similarity (OKS)~\cite{Lin2014} metrics. For PCK accuracy, we use a threshold of 0.05. In Table~\ref{tab:wheelchair_model_performance}, we also report the average PCK for each body part (columns 3--9). The Average Precision (AP) for the OKS scores (column 1) is calculated as an average over multiple thresholds from 0.50 to 0.95.

ViTPose~\cite{xu2022vitpose} variants performed best across both metrics. The knee, ankle, and head keypoints were the hardest to estimate for all models. This is likely due to occlusion when the person is not directly facing the camera, causing these keypoints to be partly obscured, while other joints are either more visible or more accurately predicted by the models despite occlusion.

While the results look promising, it is important to note that the data were collected in a laboratory environment with limited diversity in lighting and viewpoints. Future work should aim to evaluate and improve pose estimation of wheelchair users in real-world scenarios, especially for challenging body parts.

\begin{table}[tb]
  \centering
  \caption{Pose estimation performance on the wheelchair keypoints dataset. AP is computed over OKS thresholds 0.50--0.95. Per-joint PCK is reported at threshold 0.05.}
  \label{tab:wheelchair_model_performance}
  {\scriptsize
  \renewcommand{\arraystretch}{1.1}
  \begin{tabularx}{\textwidth}{@{} >{\hsize=2.5\hsize}p{0.9cm}X*{8}{S[table-format=1.3]>{\hsize=0.85\hsize}X} @{}}
    \toprule
    Model & {AP (OKS)} & {Head} & {Shoulder} & {Elbow} & {Wrist} & {Hip} & {Knee} & {Ankle} & {PCK} \\
    \midrule
    LiteHRNet-18~\cite{Yu2021litehrnet} & 74.0 & 0.833 & 0.878 & 0.811 & 0.682 & 0.919 & 0.728 & 0.714 & 0.777 \\
    LiteHRNet-30~\cite{Yu2021litehrnet} & 77.7 & 0.844 & 0.895 & 0.845 & 0.737 & 0.945 & 0.763 & 0.758 & 0.806 \\
    HRNet-w32~\cite{Sun2019hrnet} & 86.0 & 0.855 & 0.930 & 0.906 & 0.866 & 0.973 & 0.842 & 0.835 & 0.867 \\
    HRNet-w48~\cite{Sun2019hrnet} & 86.4 & 0.863 & 0.935 & 0.916 & 0.859 & 0.976 & 0.848 & 0.841 & 0.872 \\
    HRFormer-S~\cite{NEURIPS2021_3bbfdde8hrformer} & 85.2 & 0.857 & 0.919 & 0.912 & 0.848 & 0.969 & 0.840 & 0.831 & 0.862 \\
    HRFormer-B~\cite{NEURIPS2021_3bbfdde8hrformer} & 86.7 & 0.869 & 0.931 & 0.920 & 0.857 & 0.976 & 0.858 & 0.847 & 0.875 \\
    ViTPose-S~\cite{xu2022vitpose} & 85.6 & 0.863 & 0.935 & 0.910 & 0.840 & 0.967 & 0.830 & 0.816 & 0.859 \\
    ViTPose-B~\cite{xu2022vitpose} & 88.2 & 0.880 & 0.948 & 0.929 & 0.892 & 0.978 & 0.859 & 0.843 & 0.883 \\
    ViTPose-L~\cite{xu2022vitpose} & 90.9 & 0.882 & 0.956 & \textbf{0.947} & 0.935 & 0.985 & 0.888 & 0.876 & 0.905 \\
    ViTPose-H~\cite{xu2022vitpose} & \textbf{91.5} & \textbf{0.889} & \textbf{0.959} & \textbf{0.947} & \textbf{0.943} & \textbf{0.986} & \textbf{0.898} & \textbf{0.888} & \textbf{0.913} \\
    \bottomrule
  \end{tabularx}
  }
\end{table}

\section{Segment-Level Evaluation}
\label{sec:supp_segmental}

Our benchmark follows an online, per-frame protocol (Sec.4), but the frame-level boundaries also support segment-level evaluation. We report \emph{Segmental F1}@${10,25,50}$, merging each model's dense per-frame predictions into contiguous segments and matching them one-to-one against same-class ground-truth segments by intersection-over-union (IoU).

\Cref{tab:segmental_f1} reports Segmental F1 for a representative model per modality family. The results mirror the per-frame findings: the skeleton model (CNN1D) is most robust under domain shift, the frozen RGB model (VideoMAE) degrades sharply on the non-lab split, and fusion recovers much of this gap while achieving the strongest wheelchair performance.

\begin{table}[tb]
  \centering
  \caption{Segmental F1 (\%) at IoU thresholds $\{0.10, 0.25, 0.50\}$ on the in-lab, non-lab (OOD), and wheelchair test sets, for one representative model per modality family. CNN1D: 2D skeleton; VideoMAE: frozen RGB; VideoMAE+CNN1D: feature-concatenation fusion.}
  \label{tab:segmental_f1}
  \setlength{\tabcolsep}{6pt}
  {\footnotesize
  \begin{tabular}{@{}lccc@{}}
    \toprule
    & In-Lab & Non-Lab & Wheelchair \\
    Model & @10 / @25 / @50 & @10 / @25 / @50 & @10 / @25 / @50 \\
    \midrule
    CNN1D          & 79.5 / 76.5 / 64.4 & 68.8 / 64.5 / 49.1 & 61.7 / 56.8 / 41.2 \\
    VideoMAE       & 65.5 / 61.7 / 46.8 & 45.6 / 39.6 / 24.2 & 61.8 / 57.1 / 41.9 \\
    VideoMAE+CNN1D & 76.1 / 73.5 / 61.0 & 58.0 / 52.1 / 35.0 & 72.8 / 69.1 / 55.1 \\
    \bottomrule
  \end{tabular}
  }
\end{table}

\section{Annotation Details}
\label{sec:supp_annotation}

\begin{figure}[tb]
    \centering

    \begin{subfigure}[b]{1\linewidth}
        \centering
        \includegraphics[width=1\linewidth]{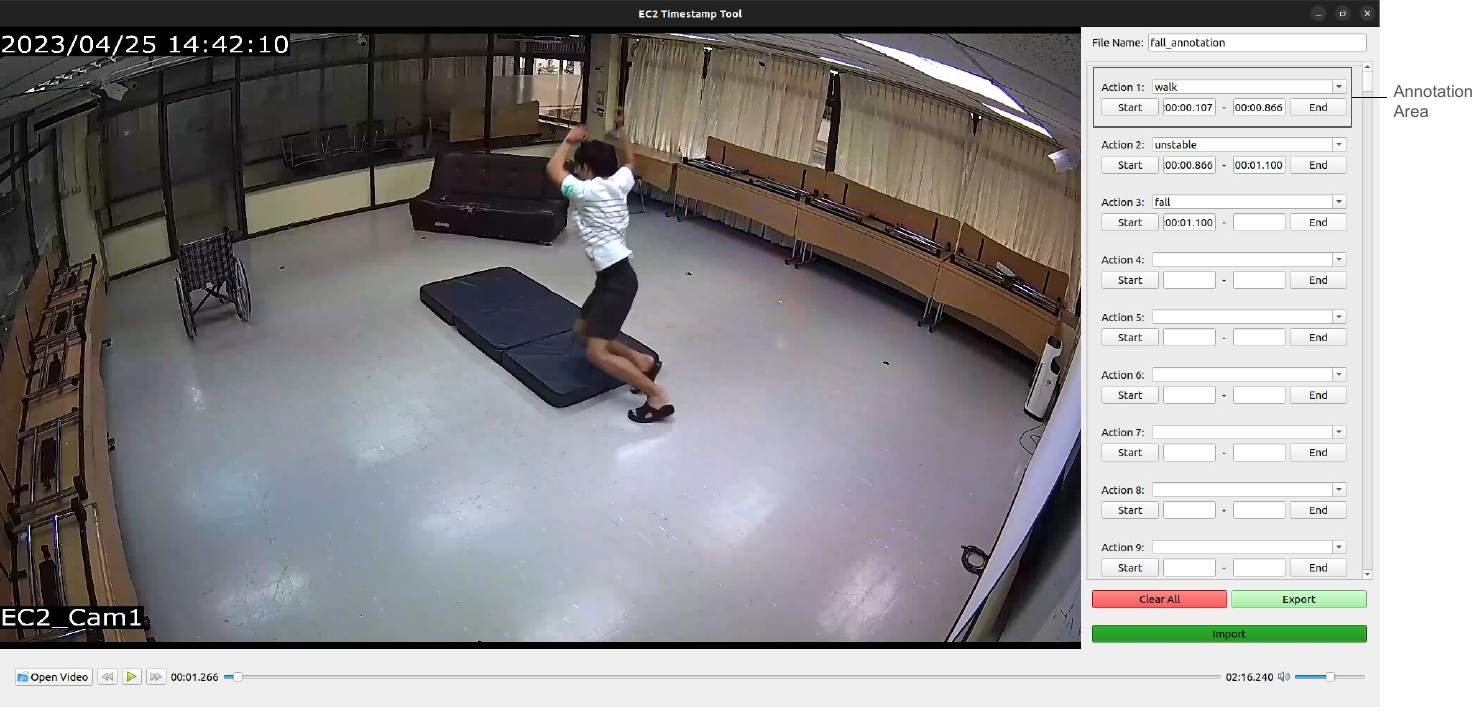}
        \caption{Annotation Tool}
        \label{fig:timestamp-tool}
    \end{subfigure}

    \begin{subfigure}[b]{1\linewidth}
        \centering
        \includegraphics[width=1\linewidth]{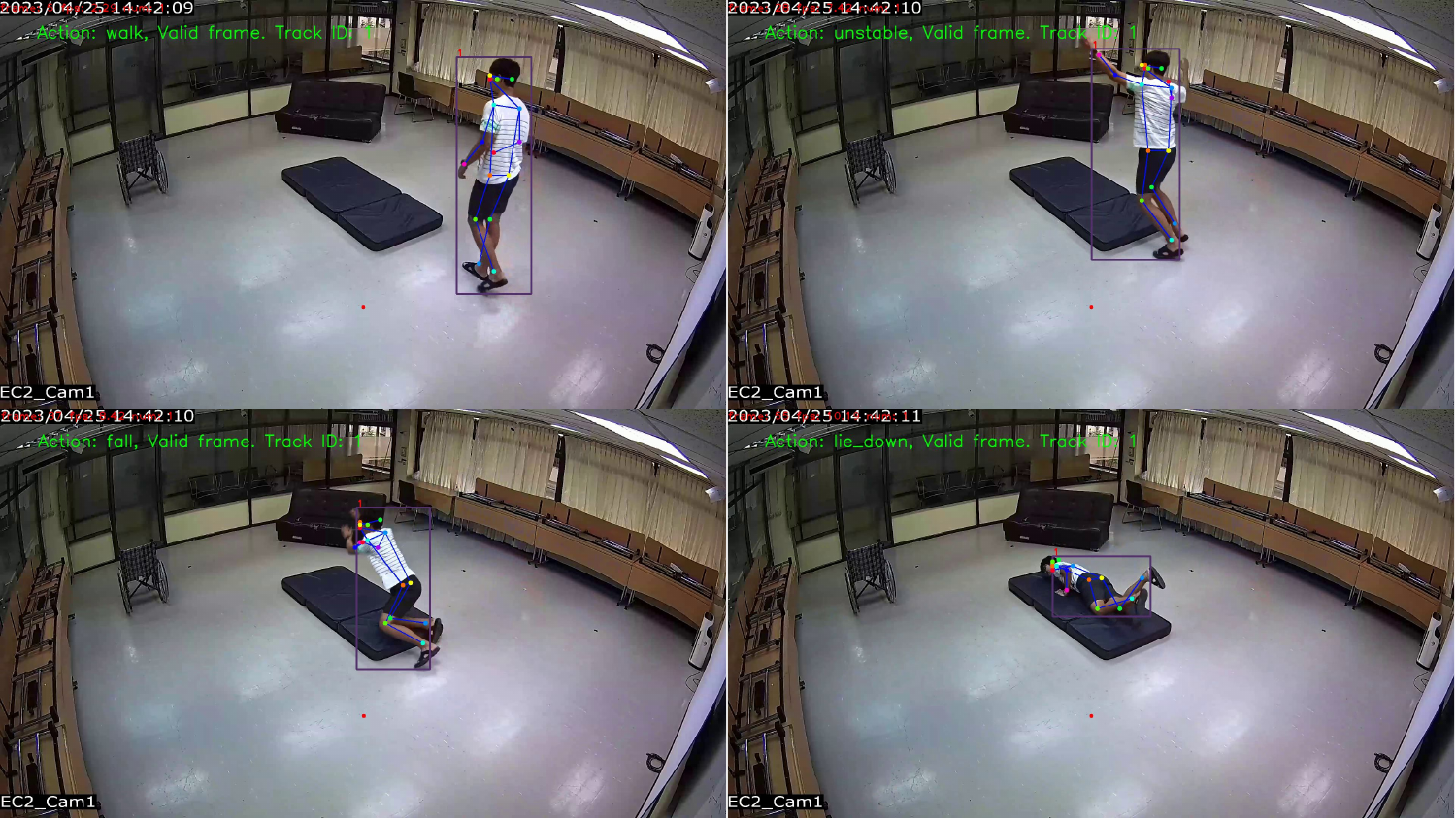}
        \caption{Ground Truth Visualization}
        \label{fig:gt-visualize}
    \end{subfigure}

    \caption{The annotation tool used by our annotators (a) and visualization of the labels overlaid on the video frames (b).}
    \label{fig:combined}
\end{figure}

Our annotators used the tool shown in Fig.~\ref{fig:timestamp-tool} to label the actions. The annotation area allows selecting from a predefined list of actions or manually entering a new action. Annotators watch the video, pausing, playing, and rewinding as needed, and use the ``Start'' and ``End'' buttons to record the current video timestamp into the corresponding text boxes. Once the entire video is annotated, it is exported to a CSV file using the ``Export'' button.

We visualized keypoint results and annotation labels (Fig.~\ref{fig:gt-visualize}) by overlaying them on videos. This keypoint visualization helped us fine-tune our pose extraction process, such as to filter out irrelevant pose extractions. The label visualization was used to identify and correct any errors in the annotations.

\section{Action Set}
\label{sec:supp_actions}

\subsection{Micro and Macro Actions}

Tables~\ref{tab:activity_instances} and~\ref{tab:wheelchair_activity_instances} show the total number of labeled instances of micro actions from the non-wheelchair and wheelchair datasets, along with the macro actions they are mapped to.

\begin{table}[tb]
  \caption{Non-wheelchair action instances. ``Instances'' is the number of labeled segments; ``All-Instances'' includes all camera views.}\label{tab:activity_instances}
  \centering
  \setlength{\tabcolsep}{5pt}
  {\small
\begin{tabular}{@{}llll@{}}
    \toprule
    Micro Action & Instances & All-Instances & Macro Action \\
    \midrule
    getting up & 1338 & 10134 & getting up \\
    walk & 1000 & 7544 & walk\\
    bend & 723 & 5625 & bend\\
    unstable & 545 & 4014 & unstable\\
    fall & 537 & 4005 & fall\\
    stand & 524 & 4149 & stand\\
    lie down & 489 & 3625 & lie down\\
    sit & 356 & 2656 & sit\\
    walk abnormal & 314 & 2254 & walk\\
    sitting down & 310 & 2449 & sitting down\\
    sit floor & 288 & 2274 & sit\\
    bend getting up & 200 & 1579 & getting up\\
    stand checktime & 98 & 784 & stand activity\\
    stand clap & 97 & 776 & stand activity\\
    stand call & 95 & 760 & stand activity\\
    stand wave & 95 & 760 & stand activity\\
    stand point & 94 & 752 & stand activity\\
    lying down & 78 & 620 & lying down\\
    sit clap & 70 & 560 & sit activity\\
    sit call & 69 & 552 & sit activity\\
    sit wave & 69 & 552 & sit activity\\
    sit checktime & 68 & 544 & sit activity\\
    sit point & 67 & 536 & sit activity\\
    reach & 55 & 440 & reach\\
    run & 53 & 424 & run\\
    stand exercise & 47 & 376 & stand activity\\
    walk walker & 29 & 232 & walk\\
    jump & 27 & 216 & jump\\
    walk cane & 23 & 184 & walk\\
    stand mop & 21 & 168 & stand activity\\
    stand complex & 15 & 120 & stand\\
    bend exercise & 14 & 112 & bend\\
    walk cane tremor & 8 & 64 & walk\\
    walk walker tremor & 6 & 48 & walk\\
  \bottomrule
  \end{tabular}
  }
\end{table}

\begin{table}[tb]
  \caption{Wheelchair action instances. ``Instances'' is the number of labeled segments; ``All-Instances'' includes all camera views.}\label{tab:wheelchair_activity_instances}
  \centering
  \setlength{\tabcolsep}{5pt}
  {\small
\begin{tabular}{@{}llll@{}}
    \toprule
    Micro Action & Instances & All-Instances & Macro Action \\
    \midrule
    sit & 476 & 3573 & sit\\
    getting up & 269 & 2003 & getting up\\
    adjust posture & 133 & 980 & sit\\
    prepare transfer & 118 & 878 & prepare transfer\\
    transfer & 106 & 786 & transfer\\
    fall & 105 & 778 & fall\\
    lie down & 102 & 758 & lie down\\
    sitting down & 101 & 747 & sitting down\\
    sit floor & 92 & 684 & sit\\
    propel & 81 & 621 & propel\\
    pick & 77 & 581 & pick place\\
    place & 76 & 574 & pick place\\
    bend & 68 & 513 & bend\\
    adjust wheelchair & 35 & 265 & sit\\
    lying down & 24 & 173 & lying down\\
    stand & 17 & 130 & stand\\
    p sit propel & 14 & 106 & get propelled\\
    drink & 13 & 99 & sit activity\\
    call & 12 & 91 & sit activity\\
    exercise & 12 & 91 & exercise\\
    deskwork & 11 & 83 & sit activity\\
    eat & 11 & 83 & sit activity\\
    reach & 3 & 24 & pick place\\
  \bottomrule
  \end{tabular}
  }
\end{table}

\subsection{Action Descriptions}

This section outlines our definitions of the macro and micro actions from the non-wheelchair and wheelchair datasets.

\subsubsection{Non-wheelchair Dataset.}

\begin{description}
    \item[stand] Standing posture.
    \begin{itemize}
        \item[] \textbf{stand\_complex} Standing in a complex position, such as with one foot raised up.
    \end{itemize}
    \vspace{2mm}

    \item[stand\_activity] Engaging in some activity while standing.
    \begin{itemize}
        \item[] \textbf{stand\_clap} Clapping while standing.
        \item[] \textbf{stand\_checktime} Checking the time while standing.
        \item[] \textbf{stand\_call} Making a phone call while standing.
        \item[] \textbf{stand\_point} Pointing at something while standing.
        \item[] \textbf{stand\_wave} Waving while standing.
        \item[] \textbf{stand\_mop} Mopping the floor while standing.
        \item[] \textbf{stand\_exercise} Doing some exercise while standing.
    \end{itemize}
    \vspace{2mm}

    \item[sit] Sitting posture.
    \begin{itemize}
        \item[] \textbf{sit\_floor} Sitting on the floor.
        \item[] \textbf{sit\_complex} Sitting in complex and casual positions, such as with knees to the chest, legs crossed, etc.
    \end{itemize}
    \vspace{2mm}

    \item[sit\_activity] Engaging in some activity while sitting.
    \begin{itemize}
        \item[] \textbf{sit\_clap} Clapping while sitting.
        \item[] \textbf{sit\_checktime} Checking the time while sitting.
        \item[] \textbf{sit\_call} Making a phone call while sitting.
        \item[] \textbf{sit\_point} Pointing at something while sitting.
        \item[] \textbf{sit\_wave} Waving while sitting.
        \item[] \textbf{sit\_exercise} Doing some exercise while sitting.
    \end{itemize}
    \vspace{2mm}

    \item[walk] Walking.
    \begin{itemize}
        \item[] \textbf{walk\_abnormal} Walking with an abnormal pattern, such as showing signs of dizziness, feet pain, etc.
        \item[] \textbf{walk\_walker} Walking with the aid of a walker.
        \item[] \textbf{walk\_cane} Walking with the aid of a cane.
        \item[] \textbf{walk\_walker\_tremor} Walking with a walker while trembling.
        \item[] \textbf{walk\_cane\_tremor} Walking with a cane while trembling.
    \end{itemize}
    \vspace{2mm}

    \item[bend] Inclining the torso forward, ranging from a slight to significant angle.
    \begin{itemize}
        \item[] \textbf{bend\_exercise} Performing bending exercises.
    \end{itemize}
    \vspace{2mm}

    \item[getting\_up] Transitioning from sitting to standing or lying down to sitting positions.
    \begin{itemize}
        \item[] \textbf{bend\_getting\_up} Getting up after a bend. Only used in some parts of the dataset. Substituted by ``getting\_up''.
    \end{itemize}
    \vspace{2mm}

    \item[sitting\_down] Transitioning from standing to a sitting position.
    \vspace{2mm}

    \item[unstable] Struggling to maintain balance, occurs before a fall most of the time.
    \vspace{2mm}

    \item[fall] Collapsing to the floor, may include different variations.
    \vspace{2mm}

    \item[lie\_down] Lying down on a surface.
    \vspace{2mm}

    \item[lying\_down] Transitioning into a lying position.
    \vspace{2mm}

    \item[reach] Extending an arm or both arms out to reach something.
    \vspace{2mm}

    \item[run] Running.
    \vspace{2mm}

    \item[jump] Jumping into the air.
\end{description}

\subsubsection{Wheelchair Dataset.}

\begin{description}
    \item[pick\_place] Interactions involving picking up and placing objects.
    \begin{itemize}
        \item[] \textbf{pick} Picking up an object.
        \item[] \textbf{place} Placing an object down.
        \item[] \textbf{reach} Reaching out for an object.
    \end{itemize}
    \vspace{2mm}

    \item[sit] Sitting position.
    \begin{itemize}
        \item[] \textbf{sit\_floor} Sitting on the floor.
        \item[] \textbf{adjust\_posture} Adjusting body posture while sitting.
        \item[] \textbf{adjust\_wheelchair} Adjusting the wheelchair position or brakes while sitting.
    \end{itemize}
    \vspace{2mm}

    \item[sit\_activity] Engaging in some activity while sitting.
    \begin{itemize}
        \item[] \textbf{deskwork} Working at a desk.
        \item[] \textbf{eat} Eating something.
        \item[] \textbf{call} Making a phone call.
        \item[] \textbf{drink} Drinking from a cup.
    \end{itemize}
    \vspace{2mm}

    \item[get\_propelled] Sitting in the wheelchair while a different person is propelling it.
    \vspace{2mm}

    \item[propel] Self-propelling the wheelchair.
    \vspace{2mm}

    \item[bend] Inclining the torso forward while seated.
    \vspace{2mm}

    \item[getting\_up] Transitioning from sitting to standing or lying down to sitting positions.
    \vspace{2mm}

    \item[exercise] Doing some exercise while in the wheelchair.
    \vspace{2mm}

    \item[sitting\_down] Transitioning from standing to a sitting position.
    \vspace{2mm}

    \item[prepare\_transfer] Preparing to transfer from or to the wheelchair.
    \vspace{2mm}

    \item[transfer] Transferring from or to the wheelchair.
    \vspace{2mm}

    \item[fall] Falling while attempting to transfer from or to the wheelchair.
    \vspace{2mm}

    \item[lie\_down] Lying down on a surface.
    \vspace{2mm}

    \item[lying\_down] Transitioning into a lying position.
    \vspace{2mm}

    \item[stand] Standing posture.
\end{description}

\section{Action Samples}
\label{sec:supp_samples}

\begin{figure}[tb]
    \centering
    \begin{subfigure}{\linewidth}
      \includegraphics[width=\linewidth]{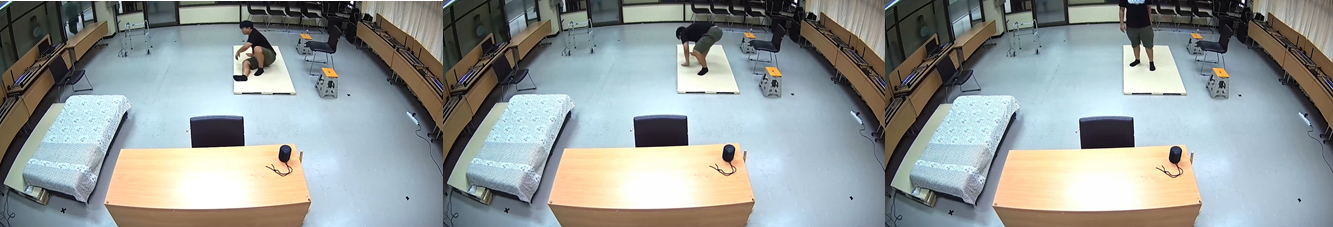}
    \end{subfigure}
    \begin{subfigure}{\linewidth}
      \includegraphics[width=\linewidth]{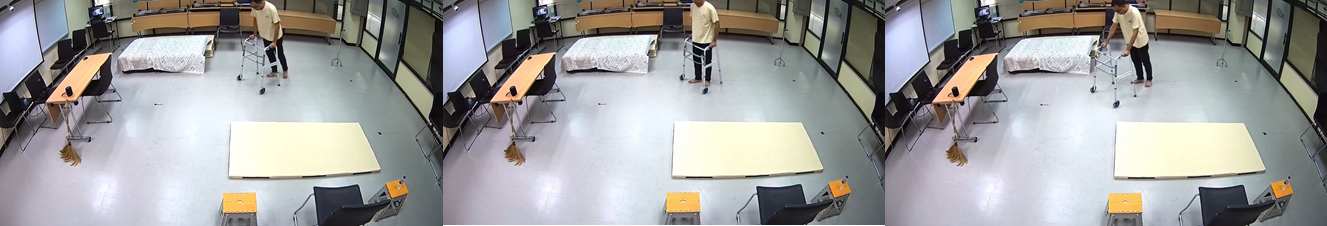}
    \end{subfigure}
    \begin{subfigure}{\linewidth}
      \includegraphics[width=\linewidth]{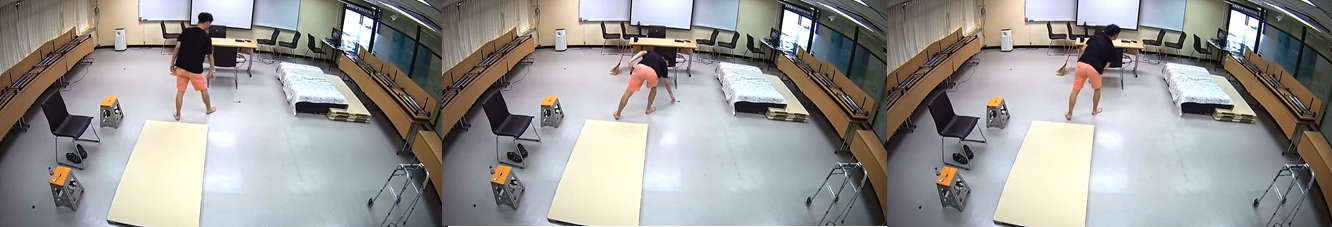}
    \end{subfigure}
    \begin{subfigure}{\linewidth}
      \includegraphics[width=\linewidth]{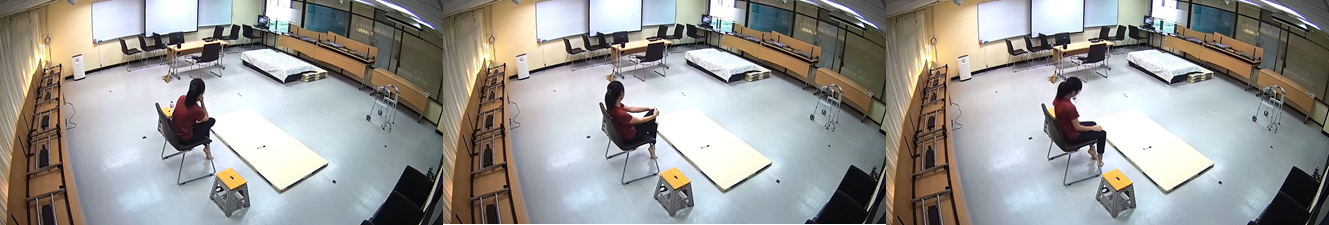}
    \end{subfigure}
    \begin{subfigure}{\linewidth}
      \includegraphics[width=\linewidth]{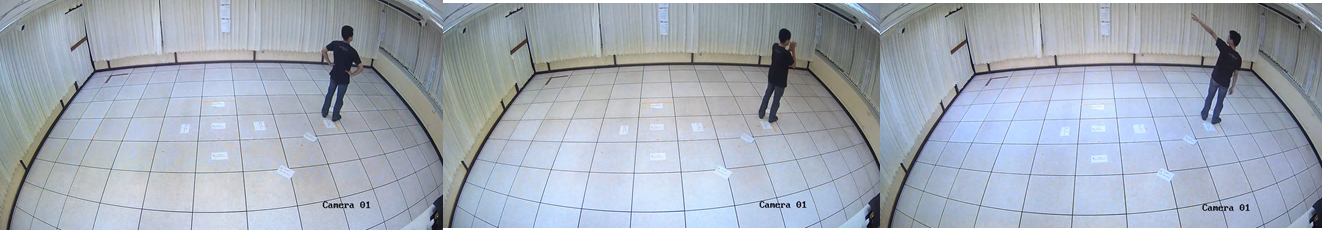}
    \end{subfigure}
    \begin{subfigure}{\linewidth}
      \includegraphics[width=\linewidth]{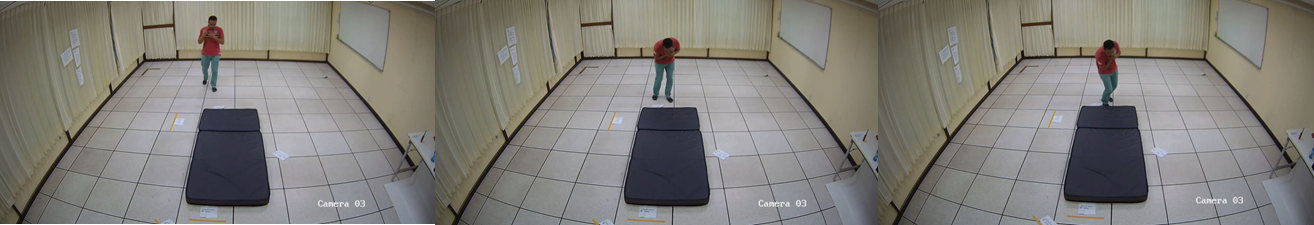}
    \end{subfigure}
    \begin{subfigure}{\linewidth}
      \includegraphics[width=\linewidth]{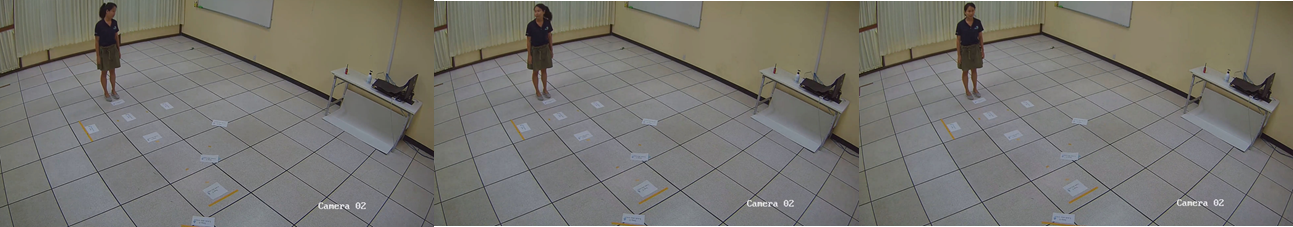}
    \end{subfigure}
    \begin{subfigure}{\linewidth}
      \includegraphics[width=\linewidth]{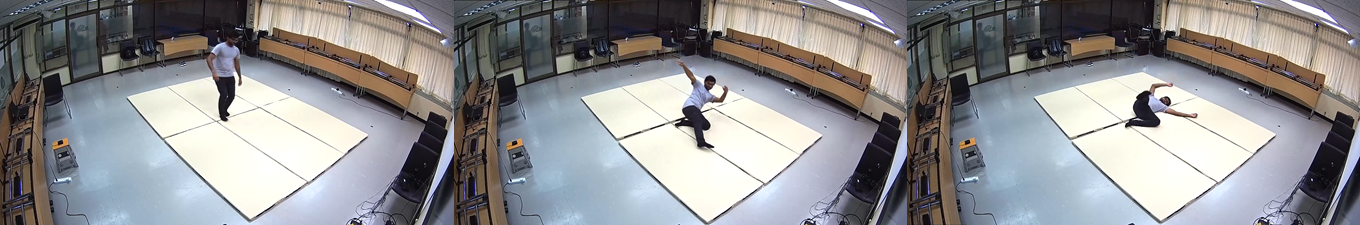}
    \end{subfigure}
  \caption{Sample actions from the non-wheelchair dataset. From top: getting\_up, walk, bend, sit, stand\_activity, unstable, stand, and fall.}
  \label{fig:nwc_action_samples_1}
\end{figure}

\begin{figure}[tb]
    \centering
    \begin{subfigure}{\linewidth}
      \includegraphics[width=\linewidth]{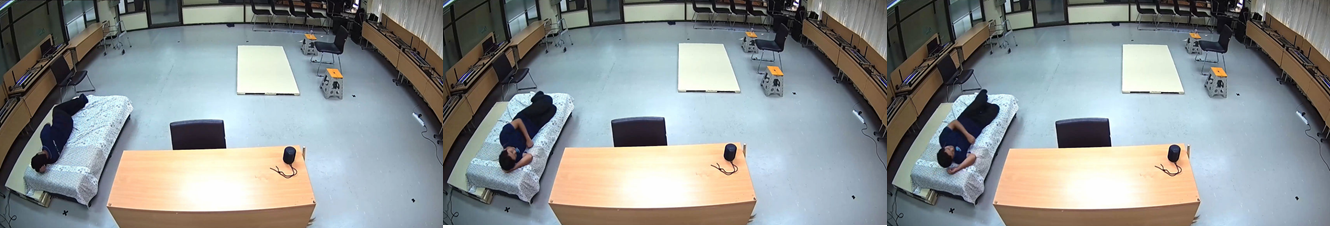}
    \end{subfigure}
    \begin{subfigure}{\linewidth}
      \includegraphics[width=\linewidth]{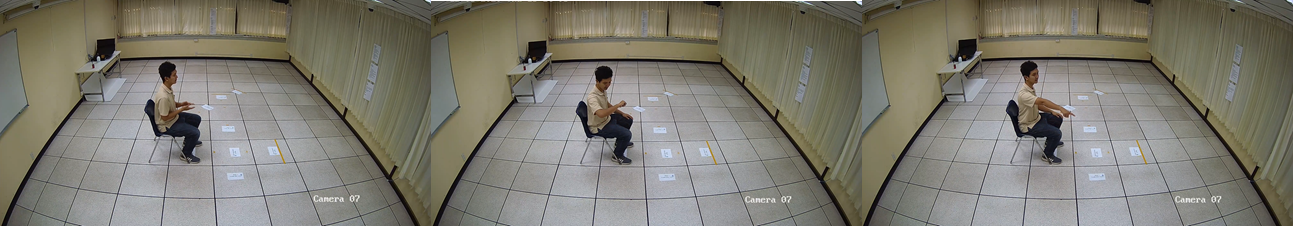}
    \end{subfigure}
    \begin{subfigure}{\linewidth}
      \includegraphics[width=\linewidth]{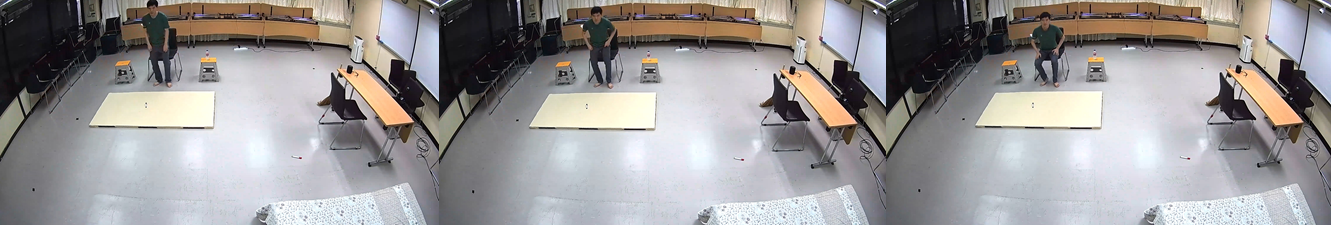}
    \end{subfigure}
    \begin{subfigure}{\linewidth}
      \includegraphics[width=\linewidth]{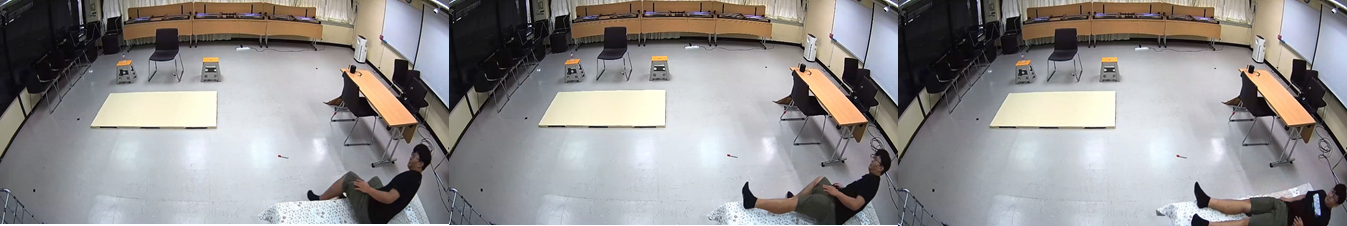}
    \end{subfigure}
    \begin{subfigure}{\linewidth}
      \includegraphics[width=\linewidth]{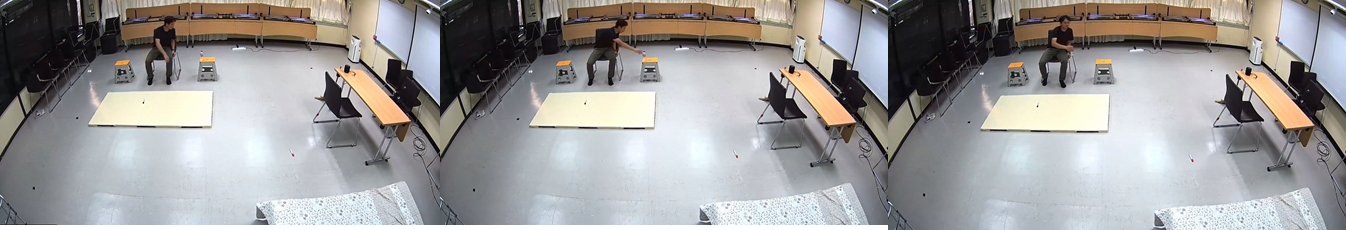}
    \end{subfigure}
    \begin{subfigure}{\linewidth}
      \includegraphics[width=\linewidth]{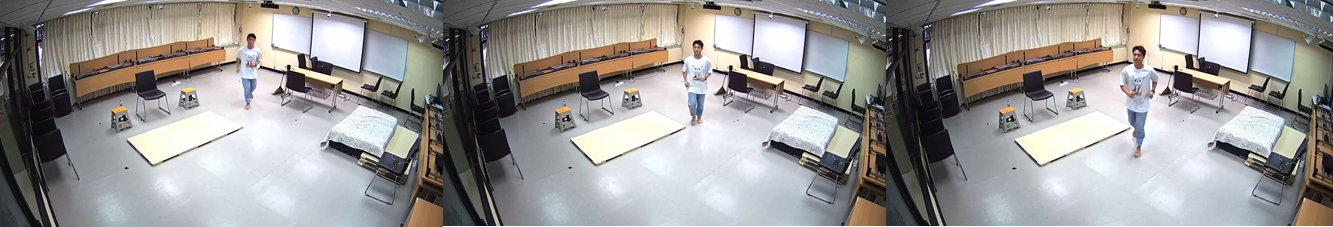}
    \end{subfigure}
    \begin{subfigure}{\linewidth}
      \includegraphics[width=\linewidth]{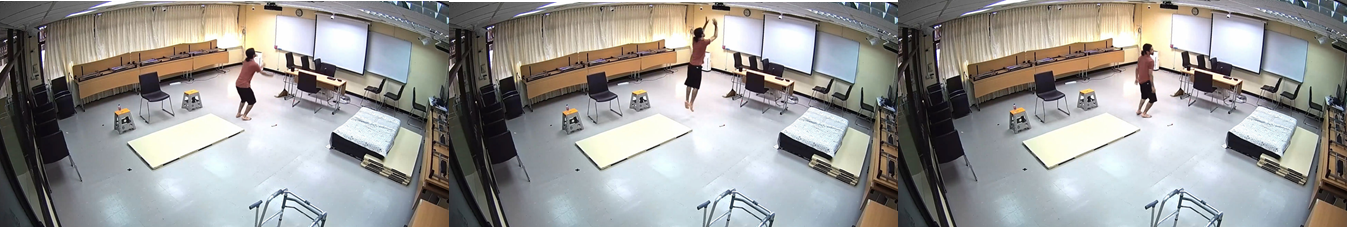}
    \end{subfigure}
  \caption{Sample actions from the non-wheelchair dataset. From top: lie\_down, sit\_activity, sitting\_down, lying\_down, reach, run, and jump.}
  \label{fig:nwc_action_samples_2}
\end{figure}

\begin{figure}[tb]
    \centering
    \begin{subfigure}{\linewidth}
      \includegraphics[width=\linewidth]{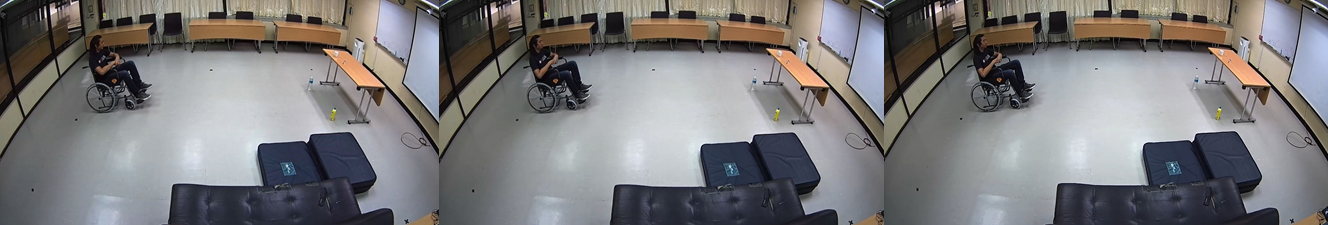}
    \end{subfigure}
    \begin{subfigure}{\linewidth}
      \includegraphics[width=\linewidth]{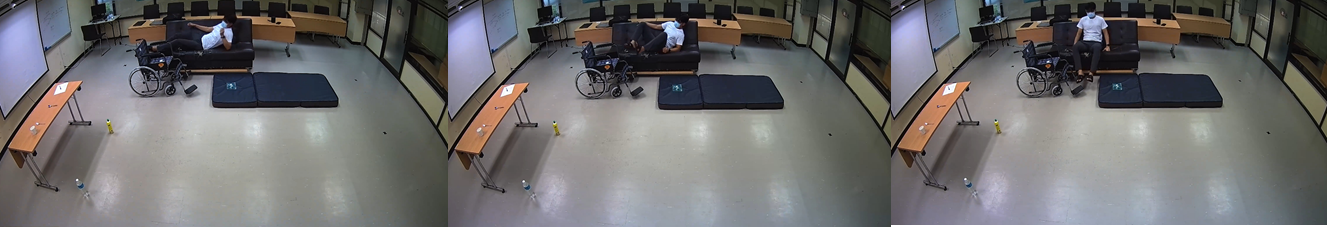}
    \end{subfigure}
    \begin{subfigure}{\linewidth}
      \includegraphics[width=\linewidth]{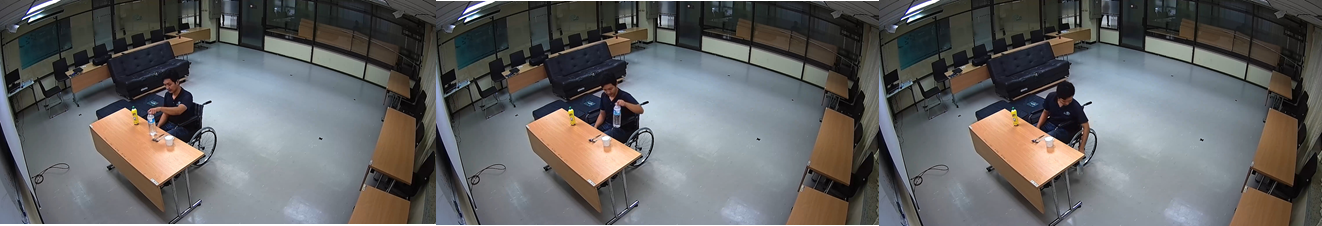}
    \end{subfigure}
    \begin{subfigure}{\linewidth}
      \includegraphics[width=\linewidth]{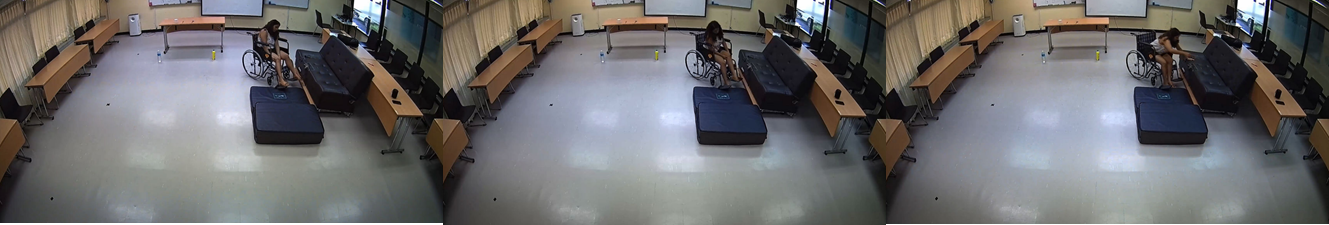}
    \end{subfigure}
    \begin{subfigure}{\linewidth}
      \includegraphics[width=\linewidth]{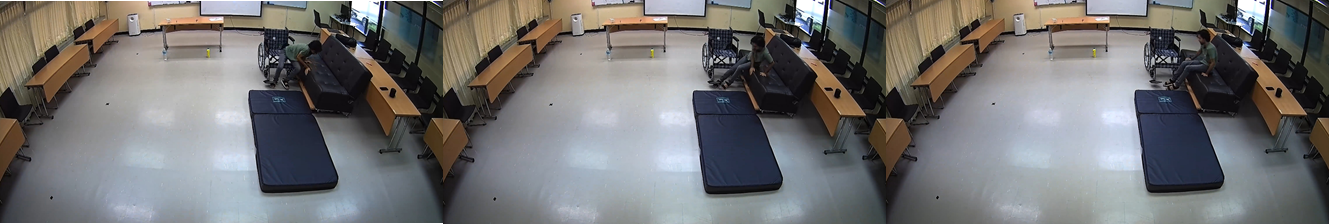}
    \end{subfigure}
    \begin{subfigure}{\linewidth}
      \includegraphics[width=\linewidth]{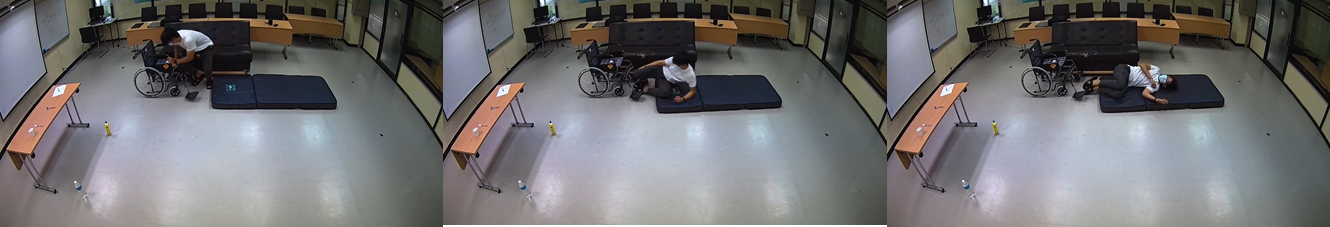}
    \end{subfigure}
    \begin{subfigure}{\linewidth}
      \includegraphics[width=\linewidth]{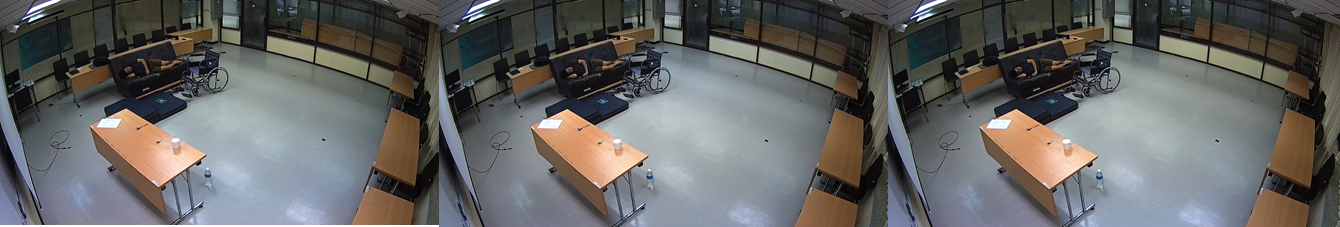}
    \end{subfigure}
    \begin{subfigure}{\linewidth}
      \includegraphics[width=\linewidth]{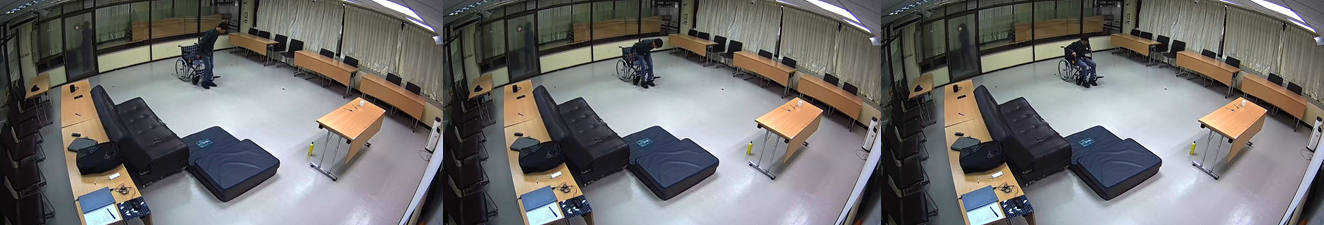}
    \end{subfigure}
  \caption{Sample actions from the wheelchair dataset. From top: sit, getting\_up, pick\_place, prepare\_transfer, transfer, fall, lie\_down, and sitting\_down.}
  \label{fig:wc_action_samples_1}
\end{figure}

\begin{figure}[tb]
    \centering
    \begin{subfigure}{\linewidth}
      \includegraphics[width=\linewidth]{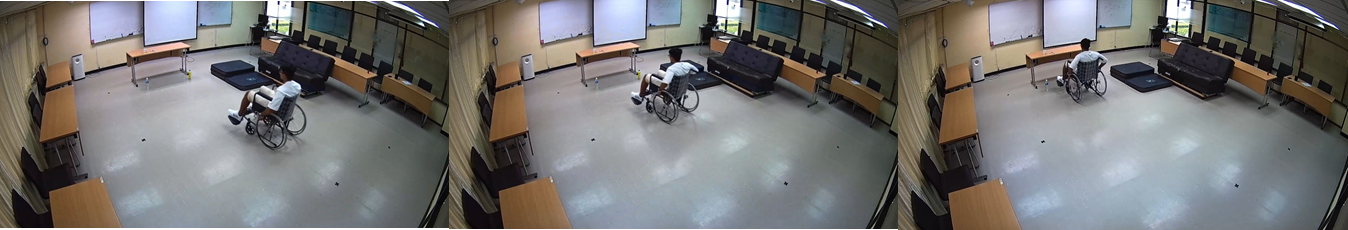}
    \end{subfigure}
    \begin{subfigure}{\linewidth}
      \includegraphics[width=\linewidth]{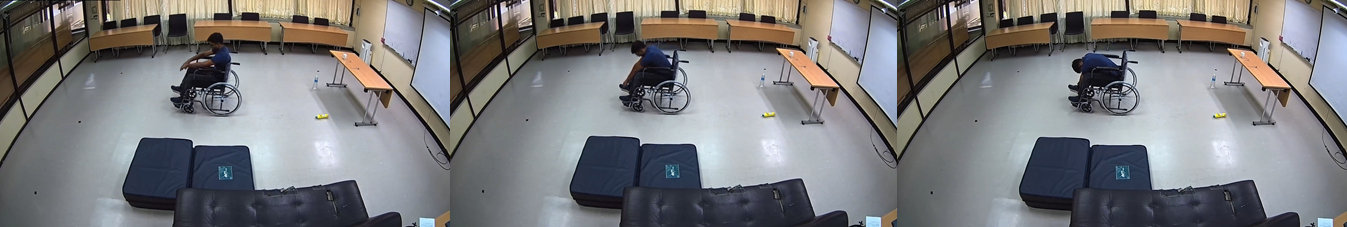}
    \end{subfigure}
    \begin{subfigure}{\linewidth}
      \includegraphics[width=\linewidth]{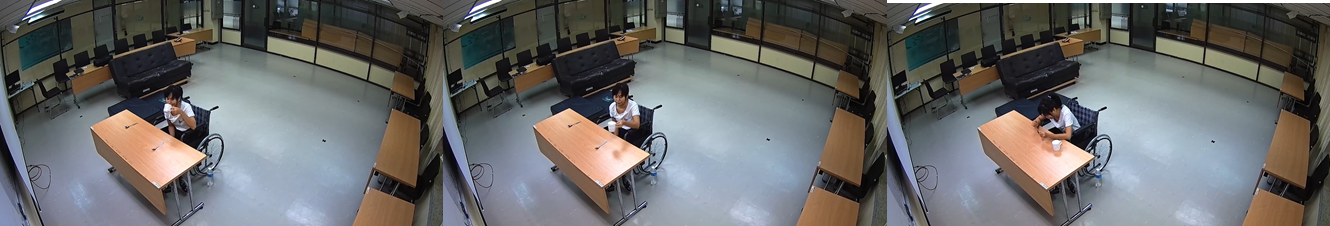}
    \end{subfigure}
    \begin{subfigure}{\linewidth}
      \includegraphics[width=\linewidth]{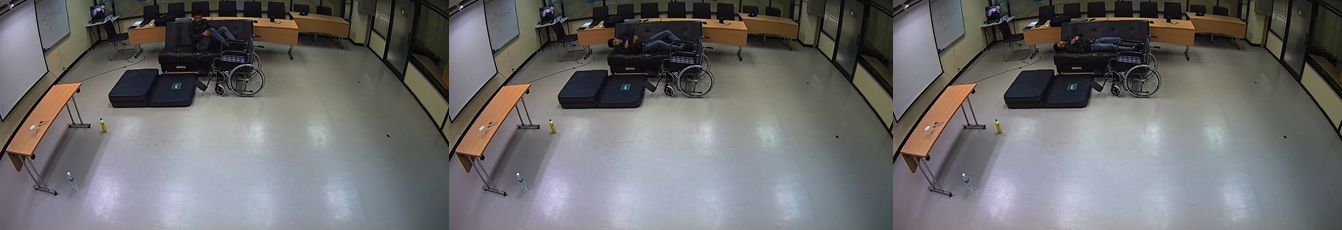}
    \end{subfigure}
    \begin{subfigure}{\linewidth}
      \includegraphics[width=\linewidth]{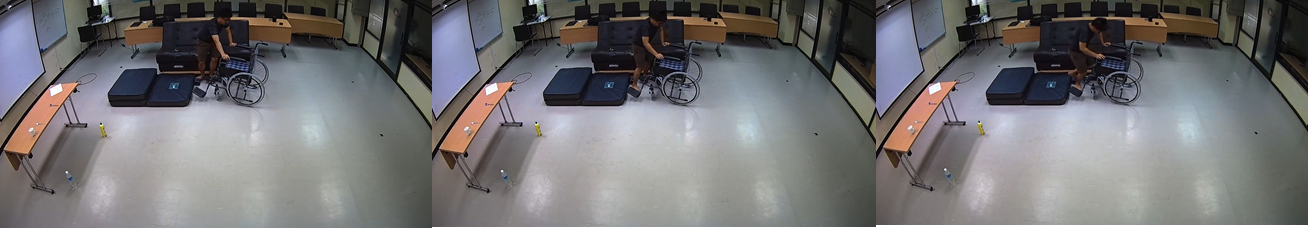}
    \end{subfigure}
    \begin{subfigure}{\linewidth}
      \includegraphics[width=\linewidth]{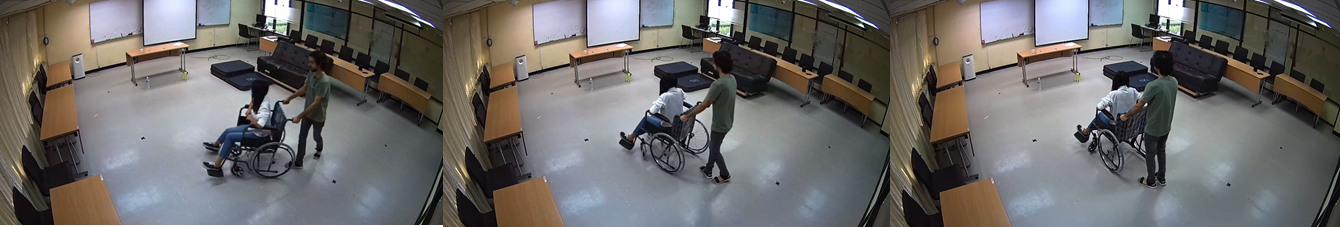}
    \end{subfigure}
    \begin{subfigure}{\linewidth}
      \includegraphics[width=\linewidth]{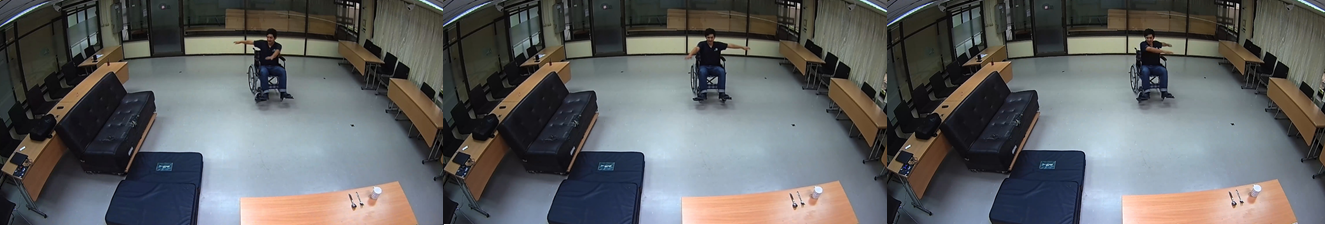}
    \end{subfigure}
  \caption{Sample actions from the wheelchair dataset. From top: propel, bend, sit\_activity, lying\_down, stand, getting\_propelled, and exercise.}
  \label{fig:wc_action_samples_2}
\end{figure}

This section presents sample actions taken from SAFER-Activities. Figures~\ref{fig:nwc_action_samples_1} and~\ref{fig:nwc_action_samples_2} show samples from the non-wheelchair dataset, while Figures~\ref{fig:wc_action_samples_1} and~\ref{fig:wc_action_samples_2} show samples from the wheelchair dataset.

\end{document}